\documentclass[lettersize,journal]{IEEEtran}
\usepackage{amsmath,amsfonts}
\usepackage{algorithm}
\usepackage{array}
\usepackage[caption=false,font=normalsize,labelfont=sf,textfont=sf]{subfig}
\usepackage{textcomp}
\usepackage{stfloats}
\usepackage{url}
\usepackage{verbatim}
\usepackage{graphicx}
\usepackage{cite}
\usepackage{booktabs}
\usepackage{lscape}
\usepackage{soul, color, xcolor}
\usepackage{multirow}
\usepackage{epstopdf}
\usepackage{algorithm}
\usepackage{algorithmicx}
\usepackage{algpseudocode}
\usepackage{float} 
\usepackage{subfloat}
\usepackage{caption}
\usepackage[english]{babel}
\usepackage{blindtext}
\usepackage{tabularx}
\usepackage{threeparttable} 
\usepackage{subcaption}
\usepackage{comment}
\usepackage{makecell}
\usepackage{booktabs}
\usepackage{array}
\usepackage{multirow}
\usepackage{hyperref}
\hypersetup{hypertex=true,
colorlinks=true,
linkcolor=red,
anchorcolor=red,
citecolor=red}

\soulregister\cite7

\soulregister\eqref7

\begin{document}

\title{An Iterative Task-Driven Framework for Resilient LiDAR Place Recognition in Adverse Weather}

\author{Xiongwei Zhao, \IEEEmembership{Graduate Student Member, IEEE}, Xieyuanli Chen, \IEEEmembership{Member, IEEE}, Xu Zhu, \IEEEmembership{Senior Member, IEEE}, Xingxiang Xie, Haojie Bai, Congcong Wen, \IEEEmembership{Member, IEEE}, Rundong Zhou and Qihao Sun


\thanks{Xiongwei Zhao, Xu Zhu and Haojie Bai are with the School of Electronic and Information Engineering, Harbin Institute of Technology (Shenzhen), Shenzhen 518071, China (e-mail: xwzhao@stu.hit.edu.cn, xuzhu@ieee.org and hjbai@stu.hit.edu.cn). (Corresponding author: Xu Zhu)}

\thanks{Xieyuanli Chen is with the College of Intelligence Science and Technology, National University of Defense Technology, Changsha 410073, China (email: xieyuanli.chen@nudt.edu.cn).}

\thanks{Xingxiang Xie is with the School of Information Communication Technology, Shenzhen Institute of Information Technology, Shenzhen 518071, China (email: xiexingxiang180@sziit.edu.cn).}

\thanks{Congcong Wen is with the Harvard AI and Robotics Lab, Harvard University, MA 02115, USA. (email: cwen2@meei.harvard.edu).}
\thanks{Qihao Sun is with the State Key Laboratory of Robotics and System, Harbin Institute of Technology, Harbin 150006, China (email: 23S008047@stu.hit.edu.cn).}
\thanks{Rundong Zhou is with the School of Electronic and Communication Engineering, Shenzhen Polytechnic University, Shenzhen 518071, China. (email: zhourd@szpu.edu.cn).}

\thanks{The datasets and code will be made publicly available at {\textcolor{red}{\textit{\url{https://github.com/Grandzxw/ITDNet}}}}}
}

\maketitle

\begin{abstract}
LiDAR place recognition (LPR) plays a vital role in autonomous navigation. 
However, existing LPR methods struggle to maintain robustness under adverse weather conditions such as rain, snow, and fog, where weather-induced noise and point cloud degradation impair LiDAR reliability and perception accuracy. To tackle these challenges, we propose an Iterative Task-Driven Framework (ITDNet), which integrates a LiDAR Data Restoration (LDR) module and a LiDAR Place Recognition (LPR) module through an iterative learning strategy. These modules are jointly trained end-to-end, with alternating optimization to enhance performance. The core rationale of ITDNet is to leverage the LDR module to recover the corrupted point clouds while preserving structural consistency with clean data, thereby improving LPR accuracy in adverse weather. Simultaneously, the LPR task provides feature pseudo-labels to guide the LDR module's training, aligning it more effectively with the LPR task. To achieve this, we first design a task-driven LPR loss and a reconstruction loss to jointly supervise the optimization of the LDR module. Furthermore, for the LDR module, we propose a Dual-Domain Mixer (DDM) block for frequency-spatial feature fusion and a Semantic-Aware Generator (SAG) block for semantic-guided restoration. In addition, for the LPR module, we introduce a Multi-Frequency Transformer (MFT) block and a Wavelet Pyramid NetVLAD (WPN) block to aggregate multi-scale, robust global descriptors. Finally, extensive experiments on Weather-KITTI, Boreas, and our proposed Weather-Apollo datasets demonstrate that, ITDNet outperforms existing LPR methods, achieving state-of-the-art performance in adverse weather.



\end{abstract}



\begin{figure}[t]
  \centering
  \includegraphics[width=\linewidth]{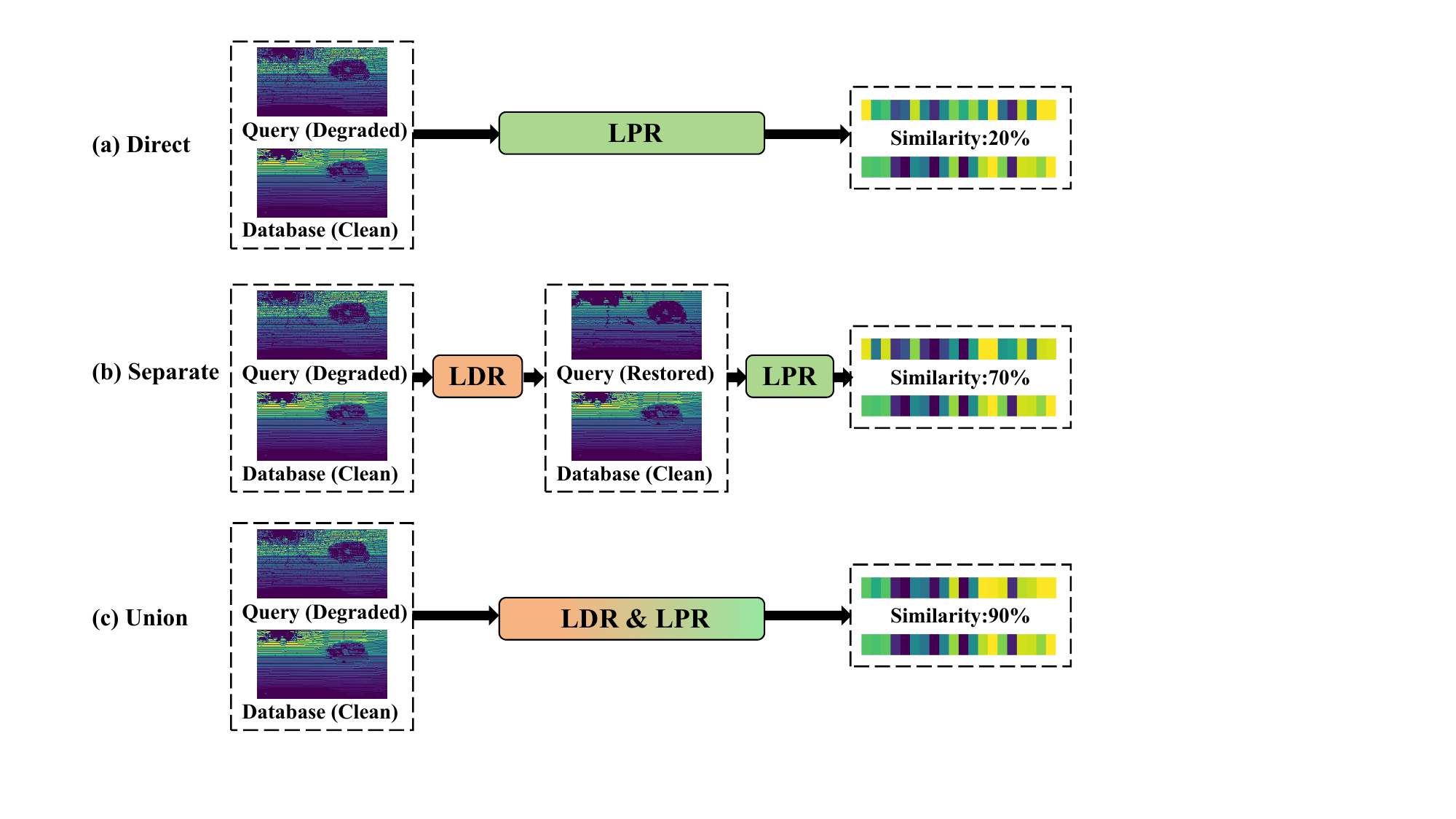}
  \caption{Three modes of LPR under adverse weather. (a) Direct Mode: The LPR model directly processes degraded query data without any preprocessing. (b) Separate Mode: The LDR module first restores the query, which is then processed by the LPR model. (c) Union Mode: The proposed iterative learning framework jointly trains the LDR and LPR modules end-to-end. 
  The proposed Union Mode achieves the highest global descriptor similarity scores.}
  \label{fig_0}
  \vspace{-0.2cm}
\end{figure}

\section{Introduction}
LiDAR Place Recognition (LPR) is a fundamental prerequisite for autonomous navigation, enabling robots to accurately localize within a global map and navigate effectively~\cite{CVTNet,mmf,mdvio,cheng2024high, liu2024apmc}. While most LPR systems exhibit robust performance under "clean" weather conditions, their reliability deteriorates significantly in adverse weather scenarios, which are common in real-world applications~\cite{adverpr,reslpr}. Conditions such as snow, rain, and fog introduce varying levels of weather-induced noise and cause partial point cloud degradation in LiDAR scans, reducing the discriminability of point cloud features and impairing the overall performance of LiDAR perception tasks~\cite{interferenc1,interferenc2,interferenc3}.
Therefore, effectively preprocessing LiDAR data while restoring it's structural consistency is essential for achieving accurate place recognition under adverse weather conditions, which has become a critical challenge for autonomous robotic navigation systems.

In recent years, researchers have developed various preprocessing techniques to enhance LiDAR perception performance in challenging environments~\cite{preprocessing1,4denoisenet, TripleMixer}. Among these, point cloud denoising is the most prevalent approach, as it effectively removes noise points, mitigating their impact on downstream tasks such as semantic segmentation and object detection~\cite{TripleMixer, Detection}. However, existing point cloud denoising methods focus solely on noise removal without restoring the structural consistency of LiDAR data. This limitation is particularly problematic for the LPR task, which relies on the similarity of global descriptors between query and database frames for accurate recognition. Meanwhile, although LiDAR data preprocessing and LPR are both essential for robust localization, they are typically treated as independent tasks~\cite{dlc-net, reslpr}. In most cases, as shown in Fig.~\ref{fig_0}(b), preprocessing methods are first applied to refine point cloud quality, and the processed data is subsequently fed into a pre-trained LPR model for inference and evaluation~\cite{reslpr, dlc-net}. However, this separate optimization overlooks the intrinsic relationship between the two tasks~\cite{zhuang2023reloc,li2023detection}. LiDAR preprocessing focuses on low-level fidelity, while LPR targets high-level scene understanding, leading to suboptimal accuracy and higher training cost. As a result, establishing a positive correlation between LiDAR data preprocessing and the LPR task remains an open challenge.



\begin{figure}[t]
  \centering
  \includegraphics[width=\linewidth]{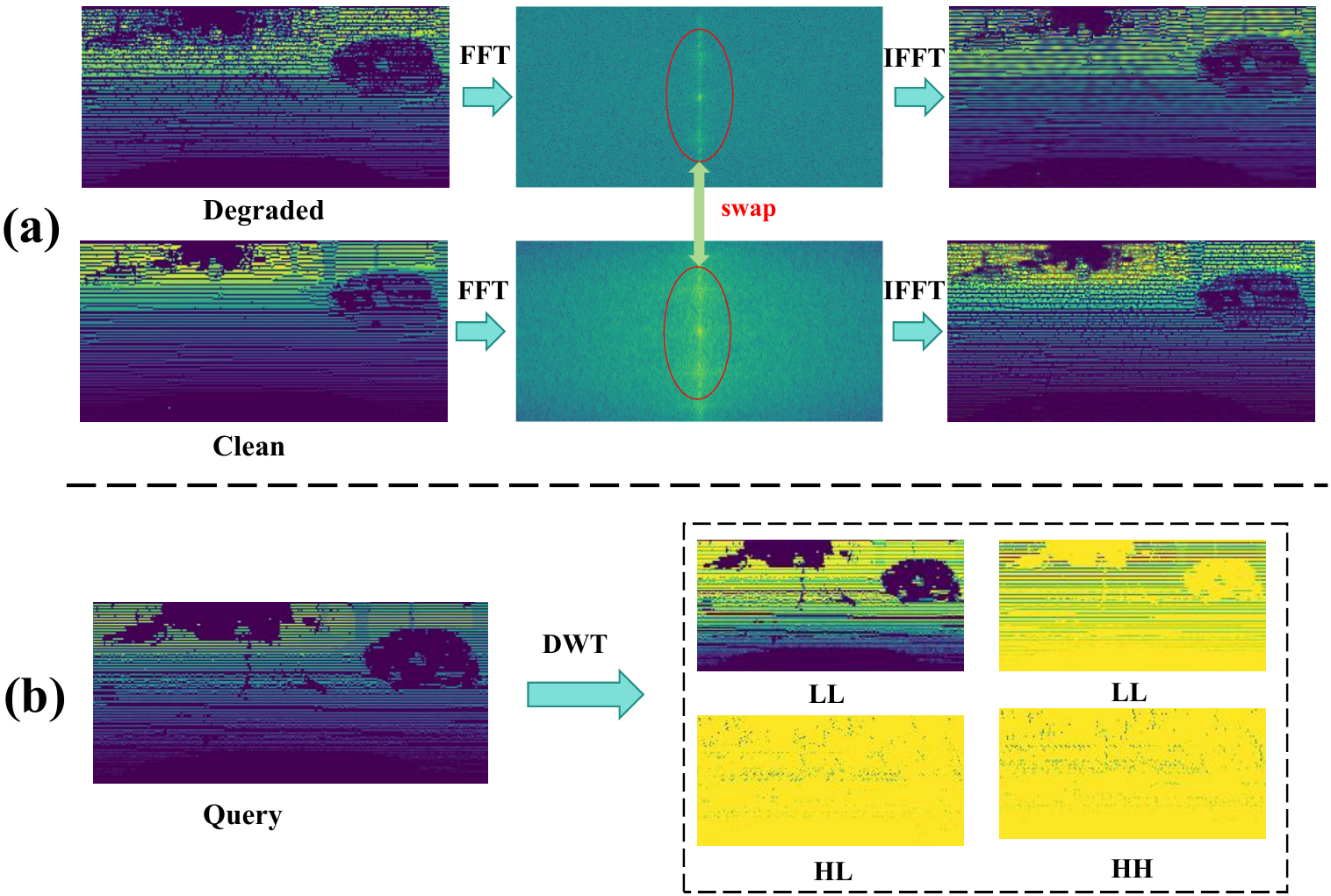}

  \caption{(a) FFT operation: The central frequency spectrum shows the FFT results of degraded and clean point clouds. After swapping the low-frequency components, the IFFT reconstructs the degraded point cloud, reducing noise and enhancing clarity. (b) DWT operation: The restored image is decomposed into four directional sub-bands (LL, LH, HL, HH) via DWT. The LL sub-band captures the global scene structure, while the LH, HL, and HH preserve fine-grained details.}
	\label{frequency}
    \vspace{-0.2cm}
\end{figure}

To address the aforementioned challenges, we propose a unified iterative learning framework that alternately trains the LPR and LDR tasks, both of which are image-based, in an end-to-end manner across alternating epochs. In our framework, the LDR module progressively generate higher-quality point cloud inputs for the downstream LPR task, while the LPR module provides pseudo-labels based on global feature representations. These pseudo-labels are then used to formulate an LPR task-driven loss, which supervises the training of the LDR module, guiding feature learning to better align the restoration process with LPR performance. Additionally, we propose frequency-domain blocks for the LDR and LPR modules to suppress noise and enhance structural feature modeling, leveraging the property that noise primarily resides in high-frequency bands while scene structures are encoded in low-frequency components. As shown in Fig.~\ref{frequency}(a), we apply the Fast Fourier Transform (FFT) in the LDR module, where low-frequency components, highlighted by the red ellipse, are concentrated near the center of the spectrum, while high-frequency components appear toward the edges. The degraded point cloud exhibits weakened energy in the low-frequency region, whereas the clean point cloud shows the opposite. By swapping the low-frequency components, the Inverse FFT (IFFT) applied to the degraded point cloud effectively suppresses noise and enhances clarity. Therefore, the FFT operator effectively isolates degradation artifacts in the frequency domain. For the LPR module, we use wavelet transform, as shown in Fig.~\ref{frequency}(b). The wavelet transform effectively isolates the scene structure in the low-frequency LL sub-band of the restored point cloud and extracts multi-scale features, which are crucial for place recognition tasks. The main contributions of this work can be summarized as follows:
\begin{itemize}
\item  We propose an end-to-end optimization framework that jointly trains the LPR and LDR modules via an iterative learning strategy. To encourage task-level interaction during training, we introduce an LPR task-driven loss, enabling the LPR module to supervise the LDR module. In return, the LDR module improves LiDAR data quality, providing better inputs for LPR.

\item We design frequency-domain blocks tailored to the LDR module. Specifically, the DDM block performs feature mixing by alternately applying frequency-domain and spatial-domain operations, effectively suppressing high-frequency noise. The SAG block further refines feature restoration through multi-scale semantic guidance.

\item For the LPR module, the MFT block enhances structural representation by applying frequency-guided window attention to multi-scale wavelet sub-bands, capturing both global structures and directional details. The WPN block then aggregates the multi-level features into a compact global descriptor for robust place recognition.

\item Extensive experiments conducted on both the real-world Boreas dataset, Weather-KITTI, and our newly introduced Weather-Apollo datasets demonstrate that the proposed ITDNet model achieves state-of-the-art performance in place recognition under adverse weather conditions. To support further research, we have open-sourced our code.

\end{itemize}


\begin{figure*}[t]
  \centering
  \captionsetup{aboveskip=2pt, belowskip=0pt}
  \includegraphics[width=\linewidth]{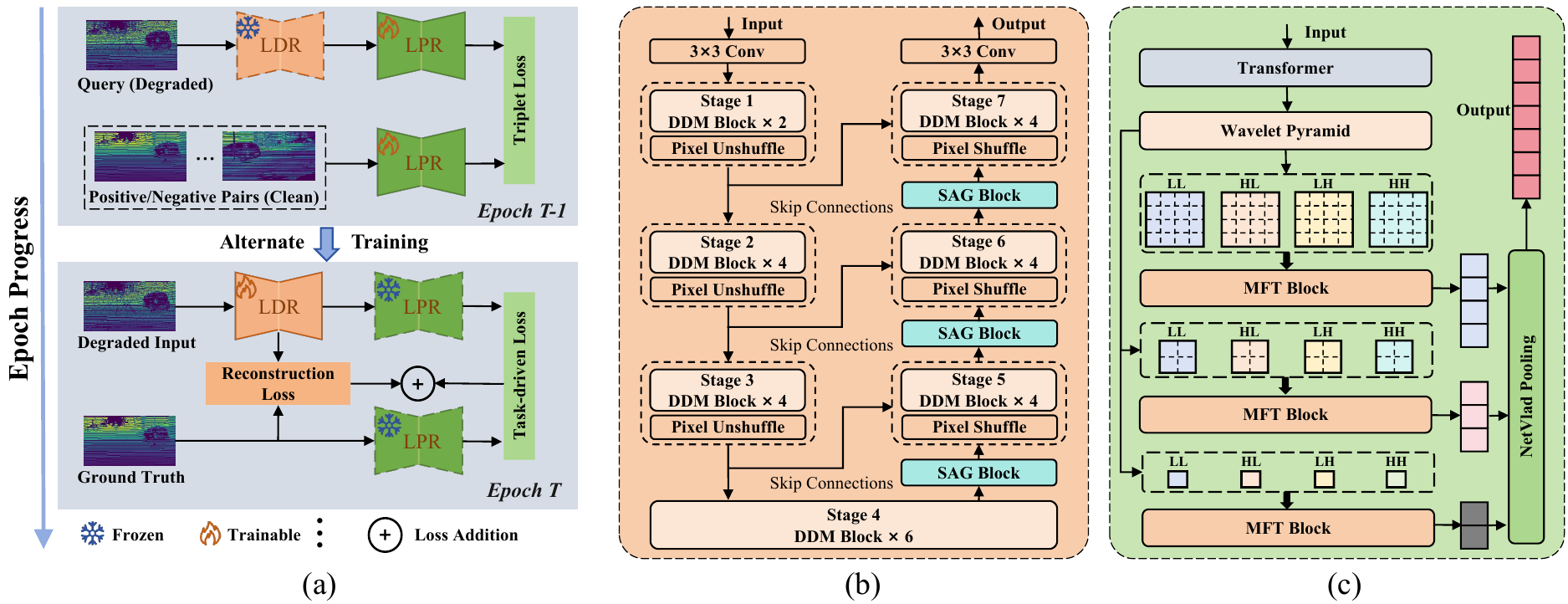}
   \caption{The Overall Architecture of ITDNet. (a) Iterative learning process of ITDNet, the LDR and LPR modules are alternately optimized across epochs. (b) Pipeline of the LDR module. The LDR module adopts a three-stage U-Net encoder-decoder framework, with DDM blocks for frequency-spatial feature modeling and SAG blocks for learning diverse noise semantics. (c) Pipeline of the LPR module. The LPR module extracts multi-scale features via a wavelet pyramid, enhances them with MFT blocks for direction-aware modeling, and aggregates them using a WPN block to produce a robust global descriptor.}
   \label{overview}
   \vspace{-0.5cm}
\end{figure*}

\section{Related Work}
\subsection{LiDAR Place Recognition Methods}
LiDAR Place Recognition (LPR) is essential for enhancing long-term navigation accuracy in simultaneous localization and mapping (SLAM) systems. Early methods focused on designing robust hand-crafted global descriptors, categorized into BEV-based \cite{sc,mmf}, discretization-based \cite{appearance}, and point-based approaches \cite{rohling2015fast,shi2023fast}. Recently, learning-based global descriptors have become dominant, divided into point-based and image-based methods. PointNetVLAD \cite{pointnetvlad} combines PointNet \cite{pointnet} and NetVLAD \cite{netvlad} for end-to-end global descriptor generation. LPS-Net \cite{lpsnet} introduces Bidirectional Perception Units (BPUs) to enhance context aggregation. OverlapTransformer \cite{ot} estimates scan overlap via range image matching using a Siamese network, while CVTNet \cite{CVTNet} fuses BEV and range image features through cross-attention. Although these methods perform well under clean weather, their robustness significantly degrades in adverse conditions.

\subsection{LiDAR Data Preprocessing Methods}
Adverse weather degrades LiDAR performance by introducing noise and causing partial point cloud corruption. Preprocessing methods mainly include statistical and deep learning approaches \cite{dreissig2023survey,surveyperception}.
Statistical methods remove noise based on geometric patterns but often fail in complex environments and are computationally intensive for large datasets \cite{yang2023learn,sor,dsor,ddior,lior}.
Deep learning methods learn noise distributions from data and can be divided into image-based and 3D-based approaches \cite{TripleMixer,4denoisenet,mobileweathernet}.
Image-based methods~\cite{wetahernet,4denoisenet} project 3D point clouds into 2D to improve real-time performance, while 3D-based methods~\cite{TripleMixer} directly process raw data to better preserve geometric structure. Although denoising improves downstream tasks such as semantic segmentation and detection, its impact on place recognition remains limited, as the task requires strong structural consistency between query and database scans. To address this,~\cite{reslpr} introduces an independent LiDAR restoration network to support place recognition. However, the method treats restoration and recognition as separate tasks without joint optimization, resulting in suboptimal performance.

\subsection{Task-Coupled Perception in Adverse Weather}
Adverse weather conditions degrade sensor data quality and negatively affect downstream perception tasks. Existing perception methods typically incorporate preprocessing modules to enhance input quality and improve perception performance. Based on the degree of task coupling, these methods can be categorized into separate learning and union learning~\cite{zhang2023perception, almalioglu2022deep}. Separate learning approaches~\cite{dlc-net, li2024ultra, sun2019convolutional, wang2023uscformer} treat preprocessing and perception as independent tasks that are optimized separately. This strategy overlooks their inherent correlation and does not guarantee that the output of preprocessing benefits the downstream perception task, often leading to suboptimal end-to-end performance. In contrast, union learning approaches\cite{zhuang2023reloc, li2023detection, liu2024oriented} jointly train and optimize preprocessing and perception within a unified framework. By designing a task-driven optimization mechanism, the relationship between low-level preprocessing and high-level recognition is effectively modeled, guiding preprocessing to better support perception. This enhances task synergy, improves end-to-end performance, and reduces redundancy\cite{feng2024bridging, wan2025collaboration}. While such union learning strategies have primarily been applied to tasks like object detection and semantic segmentation, we are the first to propose an end-to-end framework that jointly optimizes LiDAR restoration and place recognition under adverse weather, strengthening task-level interaction and robustness.




\section{Methodology}
In this section, we first present the overall architecture of the proposed ITDNet. Then, we provide a detailed description of the LiDAR Data Restoration (LDR) module and the LiDAR Place Recognition (LPR) module. Finally, we introduce the iterative training strategy used to optimize the ITDNet.
\subsection{The Overall Architecture of ITDNet}
The overall structure of ITDNet is shown in Fig.~\ref{overview}, while the alternating training process is illustrated in Fig.~\ref{overview}(a). During epoch $t-1$, the LDR module is frozen, and the model from the previous epoch processes the degraded query input. The active LPR module then performs recognition tasks on both the clean database data and the processed query input, generating a Triplet Loss to train the LPR model. In the subsequent epoch $t$, the LDR module is updated using the degraded input, while the LPR module is frozen. The LPR module, pre-trained in the prior epoch, generates global feature pseudo-labels. These pseudo-labels formulate an LPR task-driven loss, which, combined with a reconstruction loss, guides the supervised training of the LDR module. This alternating training procedure is repeated across subsequent epochs, consistently following the described optimization strategy.

Fig.~\ref{overview}(b) and Fig.~\ref{overview}(c) respectively illustrate the detailed network architectures of the LDR and LPR modules. Given a point cloud $\mathbf{P}_{0} \in \mathbb{R}^{N \times 4}$, which includes 3D coordinates and intensity, we project it into a 2D range image $\mathbf{I}_{0} \in \mathbb{R}^{H \times W \times C}$ using spherical projection \cite{ot}. Here, $H \times W$ is the spatial resolution and $C$ encodes attributes such as distance $d$ and intensity $i$. In the LDR module, the input $\mathbf{I}_{0}$ is first processed by a convolution layer to extract initial features $\mathbf{F}_{0}\in\mathbb{R}^{H\times W\times C}$. It then passes through a three-stage U-Net encoder-decoder framework, where each layer integrates a DDM block. The DDM block consists of a Frequency Mixer (FMX) layer for capturing global structures using FFT and a Spatial Mixer (SMX) layer for refining local details via depthwise convolutions and pointwise convolutions. The encoder downsamples feature maps, modeling multi-scale receptive fields and producing a compact latent representation $\mathbf{F}_\text{d} \in \mathbb{R}^{\frac{H}{8} \times \frac{W}{8} \times 8C}$. A DDM block at the bottleneck further enhances spatial correlations. In the decoder, $\mathbf{F}_\text{d}$ first passes through an SAG block, which learns noise semantic characteristics. The learned noise is integrated into $\mathbf{F}_\text{d}$ and refined through DDM blocks. The decoder progressively upsamples the features, refining spatial details, while skip connections preserve both structure and fine details, ultimately generating the restored feature $\mathbf{F}_\text{r} \in \mathbb{R}^{H \times W \times C}$. For the LPR module, the restored feature $\mathbf{F}_\text{r}$ is first processed through convolutional layers and a transformer to extract multi-scale spatial features while preserving global context. These features are further refined by the MFT block, which performs direction-aware modeling across different frequency sub-bands, capturing both local structures and long-range dependencies. Finally, the WPN block aggregates the multi-level representations across scales, producing a discriminative global descriptor.

\subsection{LiDAR Data Restoration module} 
The LDR module restores the corrupted point cloud structure from the LiDAR range image while preserving geometric and semantic consistency for downstream tasks. It employs a U-Net architecture, which consists of the DDM Block and the SAG Block. To enhance feature learning, the LDR module gradually increases kernel sizes as the network deepens, expanding the receptive field to capture multi-scale features. This hierarchical design enables fine-to-coarse feature extraction, improves contextual awareness, and enhances robustness against noise.

\begin{figure}[t]
  \centering
  \includegraphics[width=0.95\linewidth]{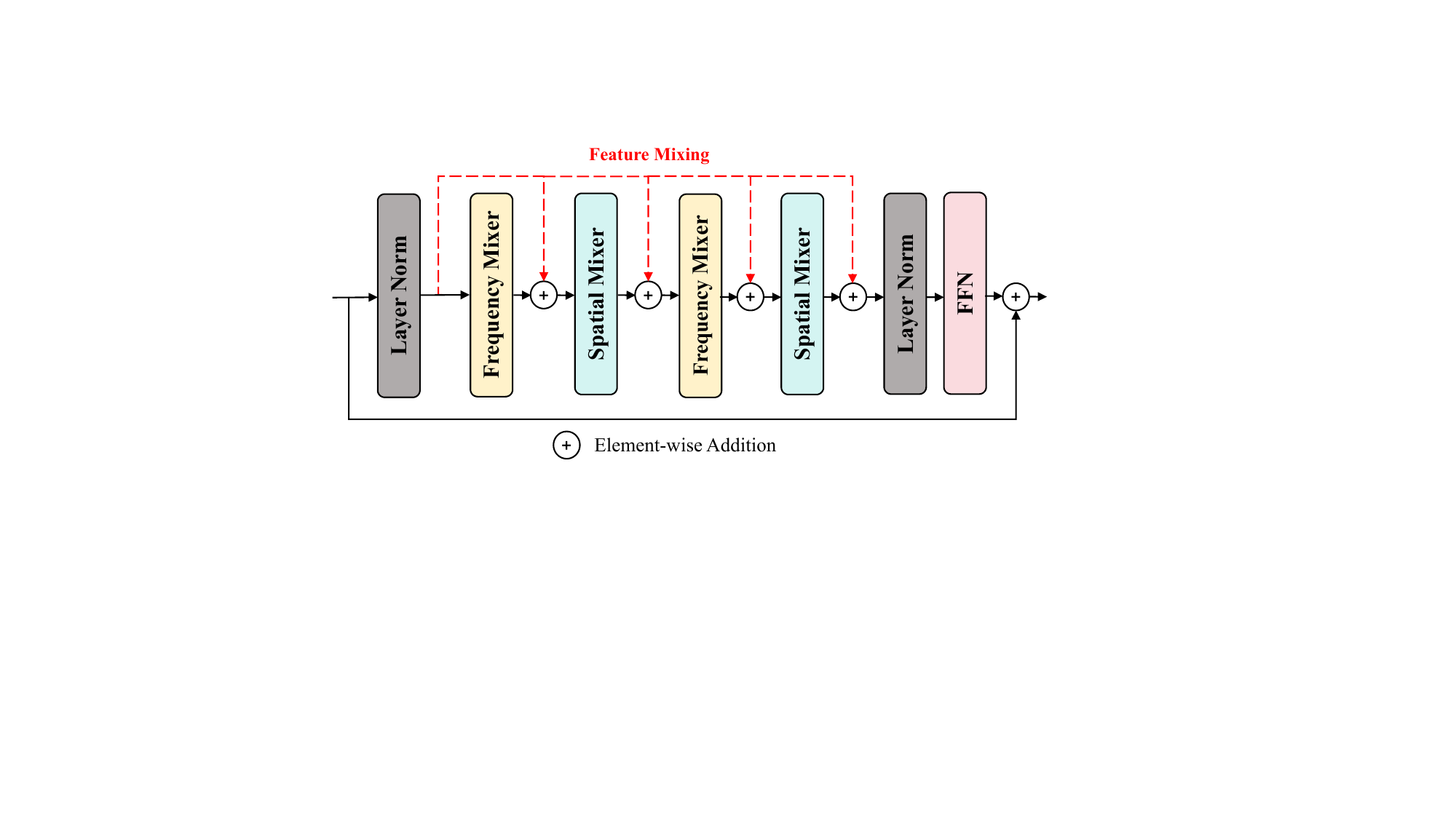}
  \caption{Details of the DDM Block, which integrates frequency and spatial mixing with progressive residual aggregation to achieve efficient global-local feature fusion.}
  \label{DDM}
  \vspace{-0.5cm}
\end{figure}

\textbf{Dual-Domain Mixer (DDM) Block}. The DDM block is designed to simultaneously capture global and local dependencies by leveraging both frequency and spatial domains in a fully convolutional manner. As shown in Fig.~\ref{frequency}(a), the FFT operator effectively separates ground-truth structures from degradations in the frequency domain, while naturally introducing global receptive fields, where perturbing a frequency component affects all spatial-domain pixels. To exploit this property, the DDM block incorporates FFT-based operations into token mixing, facilitating noise isolation and robust feature enhancement. As illustrated in Fig.~\ref{DDM}, the DDM alternates Frequency Mixers (FMX) and Spatial Mixers (SMX) across multiple stages. Each FMX model global context via FFT operations, while each SMX strengthens local feature interactions in the spatial domain. Beyond simple sequential stacking, progressive feature mixing is achieved through residual aggregation after each mixer, enabling efficient global-local information fusion while serving as a lightweight alternative to self-attention mechanisms. Given an input feature $\mathbf{F}_{0} \in \mathbb{R}^{H\times W\times C}$, the DDM block first applies Layer Normalization (LN) followed by an FMX to capture global dependencies, yielding: 
\begin{equation} 
\mathbf{F}_1 = \mathbf{F}_0 + \text{FMX}(\text{LN}(\mathbf{F}_0)). 
\end{equation} 

The intermediate feature $\mathbf{F}_1$ is then refined by an SMX to enhance local interactions in the spatial domain: 
\begin{equation} 
\mathbf{F}_2 = \mathbf{F}_1 + \text{SMX}(\mathbf{F}_1). 
\end{equation} 

This process is repeated, alternately applying FMX and SMX with residual connections: \begin{equation} 
\mathbf{F}_3 = \mathbf{F}_2 + \text{FMX}(\mathbf{F}_2), 
\end{equation} 
\begin{equation} 
\mathbf{F}_4 = \mathbf{F}_3 + \text{SMX}(\mathbf{F}_3). 
\end{equation} 

Finally, $\mathbf{F}_4$ is processed through a Layer Normalization and a Feed-Forward Network (FFN) to produce the output feature: \begin{equation} 
\mathbf{F}_{\text{out}} = \mathbf{F}_4 + \text{FFN}(\text{LN}(\mathbf{F}_4)). \end{equation}

This progressive dual-domain mixing enables the DDM block to efficiently balance global context modeling and local detail preservation while maintaining lightweight computational complexity. 

The FMX applies FFT to transform input features $\mathbf{F}_\text{0}$ into real components $\mathbf{F}{_\text{r}}$ and imaginary components $\mathbf{F}{_\text{i}}$, facilitating global information interaction. These components are refined through a convolution and a learnable selection mechanism to enhance meaningful patterns while suppressing noise. The features are then reconstructed via IFFT, enabling the subsequent SMX to effectively refine local feature interactions. 
\begin{equation} 
(\mathbf{F}{_\text{r}}, \mathbf{F}{_\text{i}}) = \text{FFT}(\mathbf{F}_\text{0}), 
\end{equation} 
\begin{equation} 
\mathbf{A} = \sigma (\text{Conv}([\mathbf{F}{_\text{r}}, \mathbf{F}{_\text{i}}])), 
\end{equation} 
\begin{equation} 
\mathbf{F}_\text{f} = \text{IFFT}(\mathbf{A} \odot [\mathbf{F}{_\text{r}}, \mathbf{F}{_\text{i}}]) + \mathbf{F}_\text{0}. 
\end{equation}
where $[\mathbf{F}{_\text{r}}, \mathbf{F}{_\text{i}}]$ represents the concatenation of the real components $\mathbf{F}{_\text{r}}$ and the imaginary components $\mathbf{F}{_\text{i}}$ of the frequency-domain features. $\sigma$ denotes the sigmoid activation function, which adaptively selects meaningful frequency patterns. $\odot$ represents element-wise multiplication, applying the learned weights 
$\mathbf{A}$ to modulate frequency responses before inverse transformation.

The SMX consists of two successive mixing layers, each comprising a depth-wise convolution (DWConv) followed by a point-wise convolution (PWConv). In each mixing layer, the DWConv applies large kernel convolutions across spatial dimensions to efficiently expand the receptive field, capturing detailed local context without significant computational overhead. Subsequently, the PWConv integrates spatially mixed features across channels, enhancing feature interaction and compactness. By stacking two identical mixing layers, SMX progressively refines spatial feature representations, effectively capturing fine-grained spatial information while preserving structural integrity and computational efficiency.
\begin{equation} 
\mathbf{F}_\text{s1} = {\delta}(\text{PWConv}({\delta}(\text{DWConv}(\mathbf{F}_\text{f})))) + \mathbf{F}_\text{f}, 
\end{equation} 
\begin{equation} 
\mathbf{F}_\text{s} = {\delta}(\text{PWConv}({\delta}(\text{DWConv}(\mathbf{F}_\text{s1})))) + \mathbf{F}_\text{s1}. 
\end{equation}
where ${\delta}$ denotes the GELU activation function, introducing non-linearity. Here, $\mathbf{F}_\text{f}$ represents the features output by the FMX, and $\mathbf{F}_\text{s1}$, $\mathbf{F}_\text{s}$ denote intermediate and final spatially refined features, respectively.

\begin{figure}[t]
  \centering
  \includegraphics[scale=0.70]{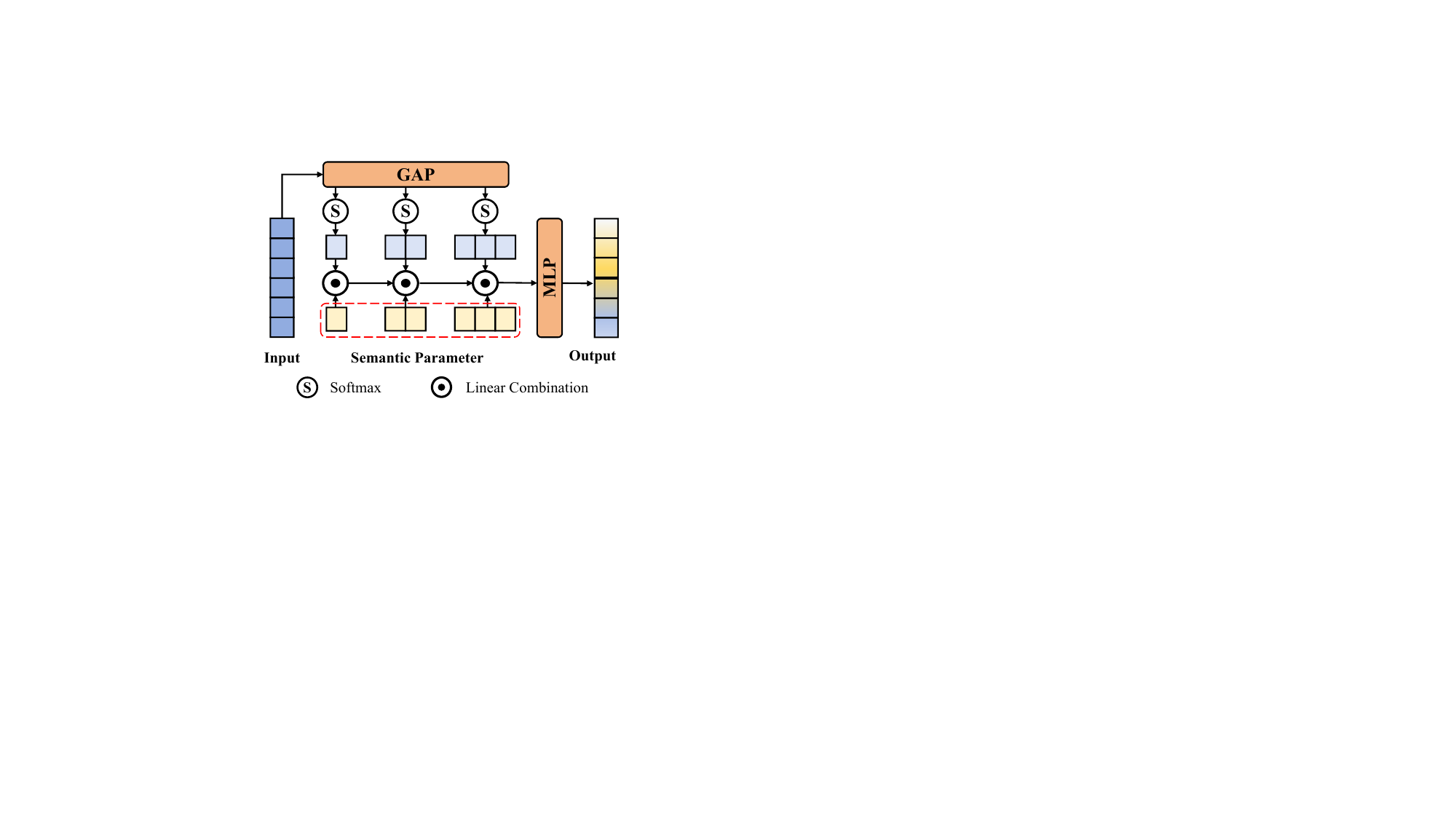}  
  \caption{Details of the SAG Block, which dynamically integrates multi-scale semantic parameters based on global context to guide feature restoration.}
  \label{SAG}
  \vspace{-0.5cm}
\end{figure}

\textbf{Semantic-Aware Generator (SAG) Block}. The SAG block leverages semantic-aware guidance to enhance feature restoration by dynamically incorporating global contextual information. As illustrated in Fig.~\ref{SAG}, the input feature first undergoes global average pooling (GAP) to produce holistic multi-scale representations. Learnable multi-scale semantic parameters are then introduced to capture structural cues at different resolutions. These parameters are dynamically weighted according to the extracted global representations, determining their contributions to the final output. By integrating multi-scale semantic guidance, SAG dynamically refines feature restoration, improving the network’s robustness in handling degraded data. The detailed process is as follows:
\begin{equation} \mathbf{e} = \text{GAP}(\mathbf{F}_\text{in}), \end{equation}
\begin{equation} \mathbf{W}_{s} = \text{Softmax}(\text{Linear}_{s}(\mathbf{e})), \quad \forall s \in \mathcal{S} \end{equation}
\begin{equation} 
\mathbf{F}' = \text{Conv}\left(\sum_{s \in \mathcal{S}}\text{Interp}_s\left(\sum_{i=1}^{N}\mathbf{W}_{i}^{s}\mathbf{F}_{i}^{s}\right)\right).
\end{equation}
where $\mathbf{F}_\text{in}$ is the input feature map, $\mathbf{e}$ is the global feature extracted via GAP, and $\mathcal{S}$ represents the set of different scales. $\mathbf{W}^{s}$ denotes the learnable scale-specific weights, while $\mathbf{F}_{i}^{s}$ is the $i$-th semantic parameter at scale $s$. $\text{Interp}_s(\cdot)$ resizes features to a common resolution before aggregation, and $\mathbf{F}'$ is the final refined feature with multi-scale semantic guidance.

\textbf{Task-driven Loss Function}. To better align the LDR module optimization with the objectives of the LPR task, we introduce an LPR-supervised task-driven loss. Specifically, in each iteration, the LPR model from the previous epoch provides global descriptors as pseudo-labels, and we enforce consistency by minimizing the KL divergence between the descriptors of restored and clean point clouds. The LDR loss $\mathcal{L}{_\text{LDR}}$ combines a reconstruction loss $\mathcal{L}_\text{REC}$ and an LPR task-driven loss $\mathcal{L}_\text{LTD}$:
\begin{equation} 
\mathcal{L}{_\text{LDR}} = \mathcal{L}{_\text{REC}} + \lambda \mathcal{L}_{\text{LTD}}, 
\end{equation}
\begin{equation}
\mathcal{L}{_\text{REC}} = \frac{1}{N} \sum_{j=1}^{N} \left( |d_j - \hat{d}_j| + |i_j - \hat{i}_j| \right),
\end{equation}
\begin{align}
\mathcal{L}_{\text{LTD}} &= D_{\text{KL}} \left( S \left( f_{\text{P}} (\hat{\mathbf{p}}_j) \right) \parallel S \left( f_{\text{P}} (\mathbf{p}_j) \right) \right) \\
&= \sum_{} S \left( f_{\text{P}} (\hat{\mathbf{p}}_j) \right) \log \frac{S \left( f_{\text{P}} (\hat{\mathbf{p}}_j) \right)}{S \left( f_{\text{P}} (\mathbf{p}_j) \right)}.
\end{align}
where the parameter $\lambda$ serves as a weighting factor to balance the contributions of the two loss components during training. In $\mathcal{L}_\text{REC}$, $d_j$ and $i_j$ represent the ground-truth distance and intensity values, while $\hat{d}_j$ and $\hat{i}_j$ are their corresponding restored predictions. $N$ denotes the total number of pixels. In $\mathcal{L}_\text{LTD}$, $S$ denotes the softmax operation, and $f_\text{P}$ refers to the LPR module obtained from the previous training epoch. $\hat{\mathbf{p}}_j$ and $\mathbf{p}_j$ denote the input features extracted from the restored and clean point clouds, respectively, and $D_{\text{KL}}$ is the Kullback–Leibler (KL) divergence, which quantifies the discrepancy between their global descriptors. By minimizing this divergence, $\mathcal{L}_\text{LTD}$ encourages the restored global descriptors to closely align with clean data. Ultimately, optimizing $\mathcal{L}_\text{LDR}$ guides the LDR module toward generating restorations that effectively support place recognition, bridging the gap between independent restoration and recognition training.

\subsection{LiDAR Place Recognition module}
The LPR module is designed to extract discriminative global descriptors from point clouds for accurate place recognition. The process begins with a transformer encoder that captures long-range dependencies. Next, a two-level wavelet decomposition is applied to generate multi-scale feature maps. Each scale is then processed by an MFT block, which performs frequency-guided attention over directional sub-bands to extract informative local features. Finally, the resulting features from all scales are aggregated through the WPN block, which fuses hierarchical frequency-aware representations into a compact and robust global descriptor.

\begin{figure}[t]
  \centering
  \includegraphics[width=\linewidth]{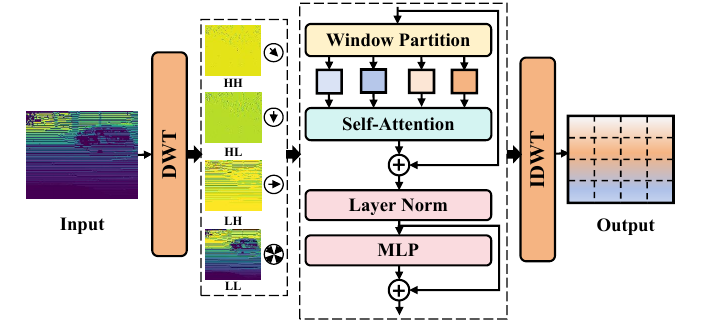}
  \caption{Details of the MFT Block. The input feature is decomposed into four directional frequency sub-bands via DWT. Each sub-band undergoes window-based self-attention to model direction-aware dependencies. The processed features are then fused and reconstructed through inverse DWT (IDWT) to obtain enhanced output representations.}
  \label{MFT}
\end{figure}

\textbf{Multi-Frequency Transformer (MFT) Block}. The MFT block enhances LiDAR-based place recognition by incorporating frequency-domain processing to capture both global structures and fine-grained details. As shown in Fig.~\ref{frequency}(b), the Discrete Wavelet Transform (DWT) decomposes features into four directional sub-bands, where the low-frequency (LL) component preserves the global scene structure, and the high-frequency (LH, HL, HH) components capture fine-grained details along specific directions. Modeling both global and directional frequency information enriches feature representations, enhancing structural discriminability and robustness under challenging conditions. However, traditional transformer attention uses fixed window shapes and sizes, limiting their ability to capture direction-specific frequency cues. To overcome this, we introduce Frequency-Guided Window Attention (FGWA), which integrates DWT with multi-window self-attention. FGWA enables direction-aware and multi-scale feature extraction, significantly improving the discriminative quality of global descriptors. Specifically, given an input feature $\mathbf{F}_{\text{0}} \in \mathbb{R}^{H\times W\times C}$, we first apply the DWT to decompose it into four sub-bands:
\begin{equation}
\mathbf{F}_\text{LL}, \mathbf{F}_\text{LH}, \mathbf{F}_\text{HL}, \mathbf{F}_\text{HH} = \text{DWT}(\mathbf{F}_{\text{0}}),
\end{equation}
where $\mathbf{F}_\text{LL}, \mathbf{F}_\text{LH}, \mathbf{F}_\text{HL}, \mathbf{F}_\text{HH} \in\mathbb{R}^{\frac{H}{2}\times \frac{H}{2}\times C}$ represent the low-frequency component, horizontal and vertical high-frequency details, and diagonal high-frequency components, respectively. To effectively capture multi-scale and directional information, we perform self-attention with four adaptively shaped windows on each frequency sub-band. Specifically, for each sub-band $\mathbf{F}_\text{s} \in \{\mathbf{F}_\text{LL}, \mathbf{F}_\text{LH}, \mathbf{F}_\text{HL}, \mathbf{F}_\text{HH}\}$, we first partition it into non-overlapping windows $[\mathbf{F}_\text{s}^1, \mathbf{F}_\text{s}^2, \dots, \mathbf{F}_\text{s}^\text{m}]$, where $m$ represents the total number of windows. Then, we apply the self-attention mechanism to each partitioned window $\mathbf{F}_\text{s}^\text{m}$ as follows:
\begin{equation}
\mathbf{A}_\text{s}^\text{m} = \text{Attention}(\mathbf{F}_\text{s}^\text{m} \mathbf{W}_\text{s}^\text{Q}, \mathbf{F}_\text{s}^\text{m} \mathbf{W}_\text{s}^\text{K}, \mathbf{F}_\text{s}^\text{m} \mathbf{W}_\text{s}^\text{V}),
\end{equation}
\begin{equation} \hat{\mathbf{F}}_\text{s} = \text{Concat}(\mathbf{A}_\text{s}^{1}, \mathbf{A}_\text{s}^{2}, ..., \mathbf{A}_\text{s}^{m}).
\end{equation}
Here, $\mathbf{W}_\text{s}^\text{Q}, \mathbf{W}_\text{s}^\text{K}, \mathbf{W}_\text{s}^\text{V} \in \mathbb{R}^{C \times d}$ are learnable projection matrices that transform the input features into query, key, and value representations. The attention-enhanced features $\mathbf{A}_\text{s}^\text{m}$ from all windows in each sub-band are concatenated to produce the refined frequency-aware feature $\hat{\mathbf{F}}_\text{s}$. This process is applied to each frequency sub-band $\mathbf{F}_\text{s}$ separately. Finally, we integrate the frequency-refined sub-band features using the Inverse Wavelet Transform (IWT) to reconstruct the enhanced spatial representation:
\begin{equation} 
\mathbf{F}{_\text{m}} = \text{IDWT}(\hat{\mathbf{F}}_\text{LL}, \hat{\mathbf{F}}_\text{LH}, \hat{\mathbf{F}}_\text{HL}, \hat{\mathbf{F}}_\text{HH}) + \mathbf{F}_{\text{0}}.
\end{equation}

By employing window attention with varying window sizes across different frequency sub-bands, our method effectively captures multi-scale directional dependencies while preserving critical structural details. This design overcomes the isotropic limitations of conventional self-attention mechanisms, as it allows adaptive feature extraction along multiple orientations. Additionally, by integrating frequency-refined components through IWT, the reconstructed spatial representation maintains both global coherence and local discriminability. Consequently, this approach enhances robustness to viewpoint variations and improves feature distinctiveness for LiDAR-based place recognition.

\textbf{Wavelet Pyramid NetVLAD (WPN) Block.} We begin by applying a two-level wavelet decomposition to extract multi-scale feature maps $\mathbf{F}_\text{0}^{i}$, where $i \in {1,2,3}$. Each feature map $\mathbf{F}_\text{0}^{i}$ is then processed by an MFT block to capture informative local patterns at different frequency levels. 
\begin{equation}
\mathbf{F}_\text{m}^{i} = \text{MFT}(\mathbf{F}_\text{0}^{i}), \quad i \in \{1,2,3\}.
\end{equation}

Then, we apply NetVLAD\cite{netvlad} encoding to each scale local features $\mathbf{F}_\text{m}^{i}$:
\begin{equation}
\mathbf{V}_\text{i} = \text{NetVLAD}(\mathbf{F}_\text{m}^{i}), \quad i \in \{1,2,3\}.
\end{equation}

Finally, we concatenate and fuse these descriptors to produce the final global representation:
\begin{equation}
\mathbf{F}_\text{w} = \text{Gating}(\text{LN}(\mathbf{W}[\mathbf{V}_\text{1}; \mathbf{V}_\text{2}; \mathbf{V}_\text{3}])),
\end{equation}
where $\mathbf{W}$ is a learnable linear projection, LN denotes LayerNorm, and Gating is the context gating mechanism. $[\cdot;\cdot]$ denotes concatenation.

\textbf{Loss Function.} 
Following prior works \cite{facenet, mmf}, we train our LPR module using the triplet loss. For each training epoch, we construct a min-batch $(x^{q}, x^{p},\left \{ x_i^{n} \right \})$ containing a query, a positive sample (a point cloud from the same location) and several negative samples (point clouds from different locations). Our objective is to ensure that the descriptor distance between the query and the positive sample is smaller than the distance between the query and any negative sample. To form the mini-batch, we select the positive sample with the closest descriptor distance from the potential positive set, ensuring stronger feature consistency. Meanwhile, negative samples are randomly selected from the potential negative set to introduce sufficient variance. The triplet loss function for the LPR module is formulated as:
\begin{equation} 
\mathcal{L}_\text{LPR}=\frac{1}{N_{\text {n}}} \sum_{i=1}^{N_{\text {n}}}\left[d\left(f(x^{q}), f\left(x^{p}\right)\right)-d\left(f(x^{q}), f\left(x_i^{n}\right)\right)+m\right]_{+},
\end{equation} 
where $d(\cdot)$ represents the Euclidean distance function, $N{_\text{n}}$ is the number of selected negative samples, and $m$ is the predefined margin enforcing the desired separation between positive and negative pairs. The notation $\left[x\right]_{+}$ denotes $\max(0, x)$, ensuring that only violations of the margin contribute to the loss.

\renewcommand{\algorithmicrequire}{\textbf{Input:}}  
\renewcommand{\algorithmicensure}{\textbf{Output:}} 
\begin{algorithm}[t]
\caption{Iterative Training Strategy for ITDNet}
\label{ITDNet}
\begin{algorithmic}[1]
\Require
$\mathbb{D}$: training dataset with  
(1) LPR triplets $(x_i^{q}, x_i^{p}, x_i^{n})$ for query, positive, and negative samples;  
(2) LDR pairs $(x_i^{n}, x_i^{c})$ for noisy input and clean ground truth.
$\mathbb{T}$: total epochs;  
$\ell_{LDR}$, $\ell_{LPR}$: learning rates for LDR and LPR;  
$\mathcal{L}_{LPR}$: loss functions for the LPR;  
$\mathcal{L}_\text{REC}$: the reconstruction loss for the LDR;
$\mathcal{L}_\text{LTD}$: the LPR task-driven loss for the LDR;
$\lambda$: task-driven loss weight.
\Ensure
$\theta_{LDR}$: trained parameters of the LDR module;
$\theta_{LPR}$: trained parameters of the LPR module.
\State \textbf{Initialize} $\theta_{LDR}$ and $\theta_{LPR}$ randomly.
\For{$epoch = 1, 2, \dots, \mathbb{T}$}
    \If{$epoch \mod 2 = 0$} \Comment{Train the LPR module}
        \For{minibatch $(x_i^{q}, x_i^{p}, x_i^{n}) \subset \mathbb{D}$}
            \State $g_{LPR} \gets \nabla_{\theta_{LPR}} \mathcal{L}_{LPR}$
            \State $\theta_{LPR} \gets AdamW(\theta_{LPR}, g_{LPR}, \ell_{LPR})$
        \EndFor
    \Else \Comment{Train the LDR module}
        \For{minibatch $(x_i^{n}, x_i^{c}) \subset \mathbb{D}$}
            \State $g_{LDR} \gets \nabla_{\theta_{LDR}} (\mathcal{L}_{REC} + \lambda \mathcal{L}_{LTD})$
            \State $\theta_{LDR} \gets AdamW(\theta_{LDR}, g_{LDR}, \ell_{LDR})$
        \EndFor
    \EndIf
\EndFor
\end{algorithmic}
\end{algorithm}

\subsection{Training Strategy of ITDNet} 
Since restoration is a low-level task aimed at recovering fine-grained pixel details, while place recognition is a high-level task focused on semantic differentiation, their optimization objectives are inherently different. Training these tasks separately can result in suboptimal performance, as the restored features may not align well with recognition requirements, reducing the overall effectiveness of both tasks. As illustrated in Algorithm \ref{ITDNet}, our approach alternates between optimizing the LDR and LPR modules. In even epochs, the LPR module is trained using triplet loss to improve place recognition accuracy. In odd epochs, the LDR module is updated with both its own reconstruction loss and an additional task-driven loss incorporating LPR supervision, ensuring that the restored features better serve the recognition task. This iterative optimization enables LDR to progressively refine its outputs, enhancing both data restoration quality and localization performance while maintaining computational efficiency.

\begin{table}[t!]
\centering
\caption{Details of the Sequences in the Boreas Dataset}
\label{boreas}
\renewcommand{\arraystretch}{1.2}
\begin{tabular}{ccc}
\toprule
\textbf{Seq ID} & \textbf{Weather Condition} & \textbf{Frame counts} \\
\midrule
2021-11-02-11-16 (Seq 00) & sun & 9568 \\
2021-03-30-14-23 (Seq 01) & sun & 11161 \\
2021-07-20-17-33 (Seq 02) & overcast, rain & 10843 \\
2020-12-01-13-26 (Seq 03) & overcast, light snowing & 9277 \\
2021-01-26-11-22 (Seq 04) & overcast, heavy snowing & 13624 \\
\bottomrule
\end{tabular}
\end{table}

\section{Experiments}
In this section, we introduce the datasets and their splits, describe the experimental setup, including implementation details and evaluation metrics, compare our method with state-of-the-art models, and present ablation studies to validate the effectiveness of each module.

\begin{table*}[h]
\centering
\setlength{\tabcolsep}{2.5pt}
\caption{\centering{Benchmarking results of LPR methods on Weather-KITTI. ITDNet-D and ITDNet-P denote the LDR and LPR modules of ITDNet, respectively.}}
\begin{tabular}{c@{\hspace{15pt}}c cccccccccccc}

\toprule

\multirow{3}{*}[-1.0ex]{\textbf{Strategy}} & \multirow{3}{*}[-1.0ex]{\textbf{Methods}} 
& \multicolumn{6}{c}{\textbf{Seq 00}} & \multicolumn{6}{c}{\textbf{Seq 07}} \\

\cmidrule(lr){3-8} \cmidrule(lr){9-14}
& & \multicolumn{2}{c}{\textbf{Snow}} & \multicolumn{2}{c}{\textbf{Fog}} & \multicolumn{2}{c}{\textbf{Rain}} 
  & \multicolumn{2}{c}{\textbf{Snow}} & \multicolumn{2}{c}{\textbf{Fog}} & \multicolumn{2}{c}{\textbf{Rain}} \\
  
\cmidrule(lr){3-4} \cmidrule(lr){5-6} \cmidrule(lr){7-8}
\cmidrule(lr){9-10} \cmidrule(lr){11-12} \cmidrule(lr){13-14}

& & \textbf{R@1↑} & \textbf{R@1\%↑} & \textbf{R@1↑} & \textbf{R@1\%↑} & \textbf{R@1↑} & \textbf{R@1\%↑}
  & \textbf{R@1↑} & \textbf{R@1\%↑} & \textbf{R@1↑} & \textbf{R@1\%↑} & \textbf{R@1↑} & \textbf{R@1\%↑} \\

\midrule

\multirow{4}{*}{{Direct}} 

& OT\cite{ot} &0.04 &0.31 &0.12 &0.46 &0.08 &0.41 &0.02 &0.53 &$-$ &0.18 &0.37 &0.79  \\

& LCDNet\cite{LCDNet} &0.08 &0.44 &0.11 &0.40 &0.14 &0.43 &\underline{0.21} &0.48 &\underline{0.22} &\underline{0.62} &0.28 &0.78 \\

& CVTNet\cite{CVTNet}  &0.12 &0.52 &0.14 &0.51 &{0.17} &\underline{0.57} &0.06 &0.54 &0.13 &0.53 &\underline{0.45} &0.66 \\

& ITDNet-P &\underline{0.15} &\underline{0.59} &\underline{0.20} &\underline{0.57} &\underline{0.22} &0.53 &0.13 &\underline{0.64} &0.14 &0.55 &0.42 &\underline{0.88}  \\

\midrule

\multirow{4}{*}{{Separate}} 

& OT\cite{ot} + ITDNet-D &0.66 &0.91 &0.75 &0.92 &0.73 &0.88  &0.48 &0.63 &0.72 &0.89  &0.69 &0.91\\

& LCDNet\cite{LCDNet} + ITDNet-D &0.69 &0.85 &0.35 &0.89 &0.56 &0.90 &\underline{0.61} &0.76  &0.41 &0.70 &0.45 &0.82\\

& CVTNet\cite{CVTNet} + ITDNet-D &0.70 &0.92 &0.77 &\underline{0.93} &0.64 &0.89 &0.49 &0.68 &0.56 &0.71 &0.51 &0.72 \\

& ITDNet-P + ITDNet-D &\underline{0.74} &\underline{0.94} &\underline{0.78} &\underline{0.93} &\underline{0.75} &\underline{0.91}  &0.59 &\underline{0.81} &\underline{0.73}  &\underline{0.95} &\underline{0.77} &\underline{0.95} \\

\midrule 
Union
& ITDNet &\textbf{0.84} &\textbf{0.95} &\textbf{0.81} &\textbf{0.95} &\textbf{0.81} &\textbf{0.93} &\textbf{0.79} & \textbf{0.97}&\textbf{0.85}  &\textbf{1.00} &\textbf{0.83} &\textbf{0.99} \\
\bottomrule
\label{lpr_weatherkitti}
\end{tabular}
\vspace{-0.5cm}
\end{table*}

\subsection{Dataset}
We conduct qualitative evaluations on three large-scale LiDAR datasets: Boreas \cite{Boreas}, the Weather-KITTI \cite{TripleMixer}, and newly introduced Weather-Apollo. These datasets cover a range of adverse weather conditions, with details provided below.

\textbf{Weather-KITTI} \cite{TripleMixer} is a large-scale synthetic LiDAR dataset for adverse weather scenarios, constructed from SemanticKITTI~\cite{Semantickitti} using a Velodyne HDL-64E sensor. For the LDR task, we use weather-degraded sequences (snow, fog, rain) from Seq 03–06 and 08–10, paired with their clean counterparts from SemanticKITTI~\cite{Semantickitti}. For the LPR task, we train on Seq 03–06 and 08–10, and evaluate on clean Seq 00 and 07 as the database set, with their weather-degraded versions serving as the query set to assess place recognition performance under adverse conditions.

\textbf{Weather-Apollo} is constructed based on the Apollo-SouthBay dataset~\cite{Apollo}, which features a Velodyne HDL-64E LiDAR and provides high-precision global poses for each scan. We select TrainData (Seq 00) with 4701 LiDAR frames and TestData (Seq 01) with 5677 LiDAR frames, both collected along the Highway237 route at different times. Following the simulation method in~\cite{TripleMixer}, we generate fog and rain conditions for Seq 00, and use the LISA simulator~\cite{LISA} to synthesize snow. To evaluate long-term place recognition performance, we use the clean Seq 01 as the database set, while the weather-degraded Seq 00 under snow, fog, and rain conditions serves as the test query set. Since Weather-Apollo shares the same LiDAR sensor model as Weather-KITTI, we directly apply the model trained on Weather-KITTI for evaluation.

\textbf{Boreas}~\cite{Boreas} provides over 350 km of driving data collected along a repeated route over one year, covering real-world seasonal and weather variations, including various levels of rain and snow. The platform is equipped with a 128-beam LiDAR and GPS/IMU sensors that offer high-precision ground-truth poses. We select five representative sequences from Boreas; all sequences are recorded along a similar route under different seasons and weather conditions, with their weather conditions summarized in Table~\ref{boreas}. For the LDR tasks, clean ground-truth scans are unavailable for degraded scans captured in adverse weather. To address this, we generate training pairs by projecting each degraded scan from Seq 02-03 onto a spatially aligned clean scan from Seq 00. Specifically, for a scan $\text{S}_{2}$ with pose $(\mathbf{R}_2, \mathbf{T}_2)$ in Seq 02, we identify the closest corresponding scan $\text{S}_{0}$ with pose $(\mathbf{R}_0, \mathbf{T}_0)$ from Seq 00, under the constraint that the spatial distance between the two poses is less than 0.01m and the angular difference is below $0.1^\circ$. Each point $\mathbf{p}_2$ in $\text{S}_2$ is then projected into the frame of $\text{S}_0$ via:
\begin{equation}
\mathbf{p}_0 = \mathbf{R}_0^{-1} \left( \mathbf{R}_2 \mathbf{p}_{2} + \mathbf{T}_2 - \mathbf{T}_0 \right).
\end{equation}

This process generates clean-degraded pairs for LDR training. While environmental changes may lead to some misalignment, we emphasize that our goal is not to recover exact point-to-point fidelity, but rather to enhance feature consistency for downstream LPR tasks. Therefore, this approximation proves sufficient for training, as evidenced by the improved place recognition performance in Table \ref{boreas_results}. For the LPR task, we train on Seq 00, use Seq 01 as the test database, and Seq 04, which contains the most severe weather degradations, as the test query set. Since each sequence includes both forward and reverse traversals, we define two evaluation settings to assess long-term place recognition performance: an \textbf{\emph{easy}} mode using only the forward traversal of the query set, and a \textbf{\emph{hard}} mode incorporating both directions to introduce greater viewpoint variation and challenge.

\subsection{Experimental Setup}
In our experiments, we first project the point clouds onto LiDAR range images. The input range images $\mathcal{I}_{0}$ are sized $64 \times 1440 \times 2$ for Weather-KITTI~\cite{TripleMixer} and Weather-Apollo, and $128 \times 1920 \times 2$ for Boreas~\cite{Boreas}, reflecting the number of LiDAR beams and the average point count per scan in each dataset. Both the LDR and LPR modules are trained for 200 epochs using the AdamW optimizer \cite{AdamW} with an initial learning rate of $10^{-4}$, decayed to $10^{-6}$ via cosine annealing ($t_{\text{max}} = 100$). Training samples are cropped to $32 \times 480$ patches with random flipping for data augmentation. To mitigate the impact of noisy pseudo-labels generated by the early-stage LPR model, the loss weight $\lambda$ for the LDR task is set to 0.01 for the first 30 epochs. After this warm-up period, $\lambda$ is increased to 0.1 to strengthen task-driven supervision. All experiments are run on an Ubuntu 18.04 system with dual Intel Xeon 8280 CPUs, four NVIDIA V100 GPUs, and 32 GB RAM.

For the LPR evaluation, a match is considered a true loop if the pose distance is less than 5 meters for the Weather-KITTI and Weather-Apollo datasets, and less than 15 meters for the Boreas dataset, excluding the previous 50 scans to avoid near-frame matches. We evaluate performance using four metrics: Recall@1 (R@1), the percentage of queries with a correct match in the top-1 result; Recall@5 (R@5), the percentage with at least one correct match in the top-5; Recall@1\% (R@1\%), the percentage with a correct match within the top 1\% of the database; F1 score (F1), which balances precision and recall for overall retrieval quality. For the LDR task, since our network is image-based and guided by the LPR objective, we evaluate restoration quality using the SSIM~\cite{ssim} metric, which assesses perceptual similarity based on structural patterns and pixel intensities. To further capture the task-driven benefits of restoration, we introduce a novel Feature Similarity Score (FSS), which measures the semantic alignment between restored and clean inputs by calculating the cosine similarity of their global descriptors obtained from the LPR model.

\begin{equation}
\text{FSS} = \frac{\mathbf{f}_{\text{r}} \cdot \mathbf{f}_{\text{c}}}{\|\mathbf{f}_{\text{r}}\|_2 \cdot \|\mathbf{f}_{\text{c}}\|_2},
\end{equation}
where $\mathbf{f}_{\text{r}}, \mathbf{f}_{\text{c}}$ denote the global descriptors extracted by the LPR model from the restored and clean inputs, respectively. A higher FSS indicates better alignment in feature space, implying greater task relevance for place recognition.

\begin{table*}[h]
\centering

\setlength{\tabcolsep}{2.5pt}
\caption{\centering{Benchmarking results of LPR methods on Weather-Apollo.}}
\begin{tabular}{c@{\hspace{13pt}}c cccc cccc cccc}

\toprule

\multirow{2}{*}{\textbf{Strategy}} & 
\multirow{2}{*}{\textbf{Method}}  & \multicolumn{4}{c}{\textbf{Snow}} & \multicolumn{4}{c}{\textbf{Fog}} & \multicolumn{4}{c}{\textbf{Rain}} \\

 \cmidrule(lr){3-6} \cmidrule(lr){7-10} \cmidrule(lr){11-14}
 
 & & \textbf{R@1↑} & \textbf{R@5↑} & \textbf{R@1\%↑}   & \textbf{F1↑}
  & \textbf{R@1↑} & \textbf{R@5↑} & \textbf{R@1\%↑}   & \textbf{F1↑}
  & \textbf{R@1↑} & \textbf{R@5↑} & \textbf{R@1\%↑}   & \textbf{F1↑} \\
\midrule

\multirow{4}{*}{{Direct}} 

&OT\cite{ot} &0.04 &0.09 &0.22 &0.08 &0.06 &0.14 &0.31 &0.12 &0.04 &0.08 &0.25 &0.09\\

&LCDNet\cite{LCDNet} &0.03 &0.06 &0.17 &0.05 &0.07 &0.15 &0.39 &0.14 &0.08 &0.14  &0.38 &0.12\\

&CVTNet\cite{CVTNet} &0.02 &0.03 &0.21 &0.04 &0.05 &{0.12} &{0.36} &0.10 &\underline{0.12} &\underline{0.23} &0.44 &\underline{0.21}
\\
&ITDNet-P &\underline{0.05} &\underline{0.11} &\underline{0.34} &\underline{0.10} &\underline{0.11} &\underline{0.23} &\underline{0.57} &\underline{0.20} &0.08 &0.18  &\underline{0.46} &0.16 \\
\midrule

\multirow{4}{*}{{Separate}} 

&OT\cite{ot} + ITDNet-D &0.43 &0.65 &0.87 &0.60 &0.41 &0.61 &0.86 &0.58  &0.23 &0.41 &0.78 &0.38\\
&LCDNet\cite{LCDNet} + ITDNet-D &0.40 &0.56 &0.81 &0.53 &0.52 &0.71 &0.87 &0.65 &0.34 &0.48 &0.76 &0.41
\\
&CVTNet\cite{CVTNet} + ITDNet-D &0.49 &0.63 &0.83 &0.66 &0.45 &0.66 &0.87 &0.62 &\underline{0.46} &\underline{0.59} &0.81 &\underline{0.63}
\\
&ITDNet-P + ITDNet-D &\underline{0.56} &\underline{0.78} &\underline{0.93} &\underline{0.72} &\underline{0.54} &\underline{0.74} &\underline{0.92} &\underline{0.70}  &0.41 &\underline{0.59} &\underline{0.89} &0.58\\
\midrule 

Union
& ITDNet &\textbf{0.66} &\textbf{0.86} &\textbf{0.97} &\textbf{0.81} &\textbf{0.67} &\textbf{0.85} &\textbf{0.96} &\textbf{0.80} &\textbf{0.69} &\textbf{0.87} &\textbf{0.97} &\textbf{0.82}
\\
\bottomrule
\label{apollo_results}
\end{tabular}
\vspace{-0.2cm}
\end{table*}

\begin{table*}[h]
\centering

\setlength{\tabcolsep}{2.5pt}
\caption{\centering{Benchmarking results of LPR methods on Boreas.}}
\begin{tabular}{c@{\hspace{13pt}}c cccc cccc}

\toprule
\multirow{2}{*}{\textbf{Strategy}} &
\multirow{2}{*}{\textbf{Method}}  & \multicolumn{4}{c}{\textbf{Seq 04 (Easy)}} & \multicolumn{4}{c}{\textbf{Seq 04 (Hard)}} \\

 \cmidrule(lr){3-6} \cmidrule(lr){7-10}
 
& & \textbf{R@1↑} & \textbf{R@5↑} & \textbf{R@1\%↑} & \textbf{F1↑}  
& \textbf{R@1↑} & \textbf{R@5↑} & \textbf{R@1\%↑} & \textbf{F1↑} \\
\midrule

\multirow{4}{*}{{Direct}} 

&OT\cite{ot} &0.16 &0.22 &0.30 &0.23 &0.10 &0.15 &0.27 &0.18 \\

&LCDNet\cite{LCDNet} &0.05 &0.14 &\underline{0.49} &0.16 &0.03 &0.11 &0.30 &0.12\\

&CVTNet\cite{CVTNet} &0.22 &0.26 &0.37 &{0.32} &\underline{0.17} &\underline{0.21} &\underline{0.34} &\underline{0.28}
\\

&ITDNet-P &\underline{0.26} &\underline{0.28} &0.39 &\underline{0.36} &\underline{0.17} & 0.20 &0.28 &0.26 \\
\midrule

\multirow{4}{*}{{Separate}} 

&OT\cite{ot} + ITDNet-D &0.35 &0.42 &0.62 &0.49 &0.23 &0.34 &0.57 &0.38 \\
&LCDNet\cite{LCDNet} + ITDNet-D &0.23 &0.31 &0.54 &0.37 &0.19 &0.25 &0.43 &0.31
\\
&CVTNet\cite{CVTNet} + ITDNet-D &0.61 &0.70 &\underline{0.82} &0.74 &0.53 &{0.60} &0.72 &0.69
\\
&ITDNet-P + ITDNet-D &\underline{0.65} &\underline{0.72} &0.79 &\underline{0.77} &\underline{0.58} &\underline{0.63} &\underline{0.75} &\underline{0.74}\\
\midrule 

Union
&ITDNet &\textbf{0.71} &\textbf{0.76} &\textbf{0.85} &\textbf{0.81} &\textbf{0.63} &\textbf{0.69} &\textbf{0.79} &\textbf{0.78}
\\

\bottomrule
\label{boreas_results}
\end{tabular}
\vspace{-0.2cm}
\end{table*}

\subsection{Evaluation for the LPR task}
In this section, we compare our proposed ITDNet with three state-of-the-art LPR methods: the image-based OverlapTransformer (OT)~\cite{ot} and CVTNet\cite{CVTNet}, and the point-based LCDNet\cite{LCDNet}. Evaluations are conducted under adverse weather conditions in the following three setting: 1) \textbf{Direct}: each LPR model, including our LPR module (ITDNet-P), is evaluated directly on degraded dataset; 2) \textbf{Separate}: each LPR model is combined with our LDR module (ITDNet-D) via separate training; 3) \textbf{Union}: our full ITDNet pipeline is trained end-to-end through joint iterative optimization of LDR and LPR modules.

\textbf{Results on Weather-KITTI:} 
We first evaluate performance on the Weather-KITTI dataset, as summarized in Table~\ref{lpr_weatherkitti}. The top section of the table shows results of the \textbf{Direct} setting: Our LPR module, ITDNet-P, consistently achieves superior performance in R@1 and R@1\% across most weather conditions, demonstrating robust resilience to degradation. In rain scenarios (Seq 00 and Seq 07), ITDNet-P remains highly competitive, closely following LCDNet \cite{LCDNet} with minimal margins of 0.04 (R@1\%) and 0.03 (R@1). Conversely, OT \cite{ot} shows significant drops, particularly under Seq 07 fog conditions. These observations underline the effectiveness of our multi-scale directional frequency modeling strategy. 


The middle section of Table~\ref{lpr_weatherkitti} reports the \textbf{Separate} setting, where each LPR model is paired with our independently trained LDR module (ITDNet-D). All methods show substantial performance gains, indicating that ITDNet-D effectively enhances LiDAR input quality and structural coherence, thereby facilitating downstream place recognition. The bottom section (\textbf{Union}) presents further improvements from our end-to-end trained ITDNet. Compared to separate training, ITDNet achieves additional average gains of 0.06 in R@1 and 0.02 in R@1\% on Seq 00, and 0.13 in R@1 and 0.08 in R@1\% on Seq 07 across all adverse weather scenarios. These results highlight the benefits of joint iterative optimization in aligning LiDAR restoration with place recognition performance.

\textbf{Results on Weather-Apollo:} 
We further evaluate our model on the proposed Weather-Apollo dataset, with results presented in Table~\ref{apollo_results}. All models trained on Weather-KITTI are directly evaluated without retraining. In the \textbf{Direct} setting, ITDNet-P consistently surpasses baseline methods under Snow and Fog conditions, highlighting its robust generalization capability. Under Rain conditions, CVTNet\cite{CVTNet} achieves slightly higher scores than ITDNet-P in terms of R@1, R@5, and F1, but the margins are minimal (0.04, 0.05, and 0.05, respectively). In the \textbf{Separate} setting, pairing each model with our LDR module consistently enhances performance, validating the general effectiveness of our restoration approach. Finally, our fully optimized ITDNet (\textbf{Union}) significantly outperforms separate training, especially in challenging rain conditions, with improvements of 0.28 (R@1), 0.28 (R@5), and 0.24 (F1). These results further reinforce the conclusion from the Weather-KITTI experiments, clearly demonstrating the efficacy of our iterative optimization strategy in tightly coupling LiDAR data restoration with place recognition, resulting in enhanced robustness across diverse and challenging weather conditions. 

\textbf{Results on Boreas:} 
Table~\ref{boreas_results} presents the evaluation results on the real-world Boreas dataset\cite{Boreas}. Compared to Weather-KITTI\cite{TripleMixer} and Weather-Apollo, performance improvements are less pronounced due to increased real-world complexity, such as varying weather intensities and dynamic environments that introduce diverse noise patterns. In the \textbf{Direct} setting, our method achieves the best performance under the \emph{easy} scenario and remains highly competitive under the \emph{hard} scenario, with only a small margin behind CVTNet~\cite{CVTNet}. In the \textbf{Separate} setting, LCDNet~\cite{LCDNet} shows limited gains when combined with our LDR module (ITDNet-D), likely due to its reliance on raw 3D geometry, which benefits less from image-based restoration. In contrast, image-based methods (CVTNet~\cite{CVTNet}, OT~\cite{ot}, and ITDNet-P) consistently see greater improvements, underscoring the compatibility of our restoration approach with range image inputs. Finally, our fully integrated ITDNet model (\textbf{Union}) achieves the highest performance, particularly under the challenging "hard" evaluation setting. Compared to separate training, joint iterative training yields moderate yet consistent improvements across all metrics (0.05 in R@1, 0.04 in R@5, 0.03 in R@1\%, and 0.04 in F1). These findings highlight that, despite the constraints imposed by real-world complexity, our joint optimization strategy effectively aligns restoration with place recognition, providing practical performance enhancements.

\begin{figure*}[t]
  \centering
  \includegraphics[scale=0.33]{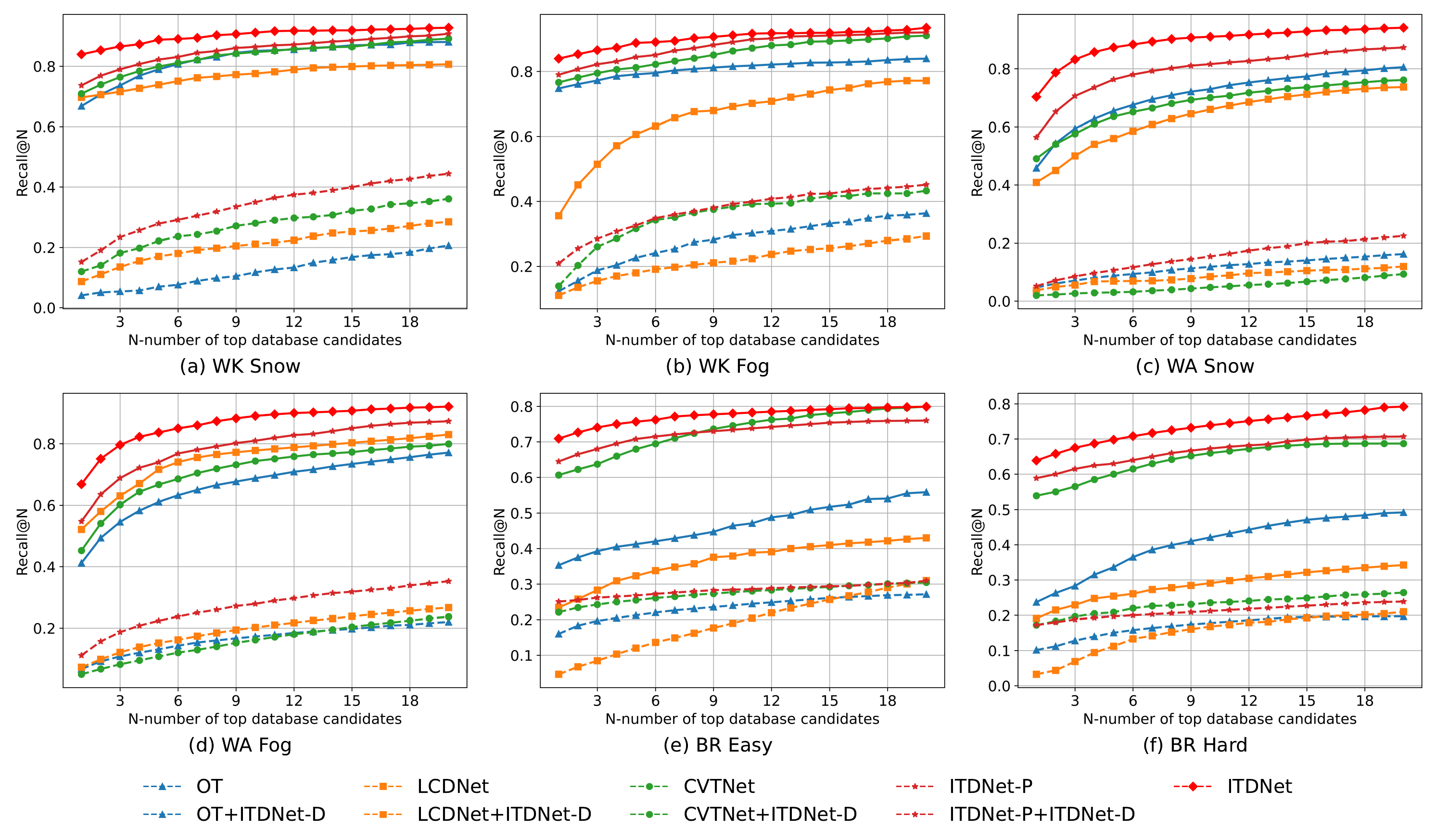}  
  \caption{Recall@N curves on the adverse weather datasets. "WK" denotes Weather-KITTI, "WA" denotes Weather-Apollo, and "BR" denotes the Boreas dataset.
"+ITDNet-D" represents each LPR model separately trained with the proposed ITDNet-D module. It can be observed that our full ITDNet achieves consistently superior performance across all Recall@N metrics.}
  \label{prvis}
  \vspace{-0.5cm}
\end{figure*}

\begin{figure*}[t]
  \centering
  \captionsetup{aboveskip=2pt, belowskip=0pt}
  \includegraphics[width=\linewidth]{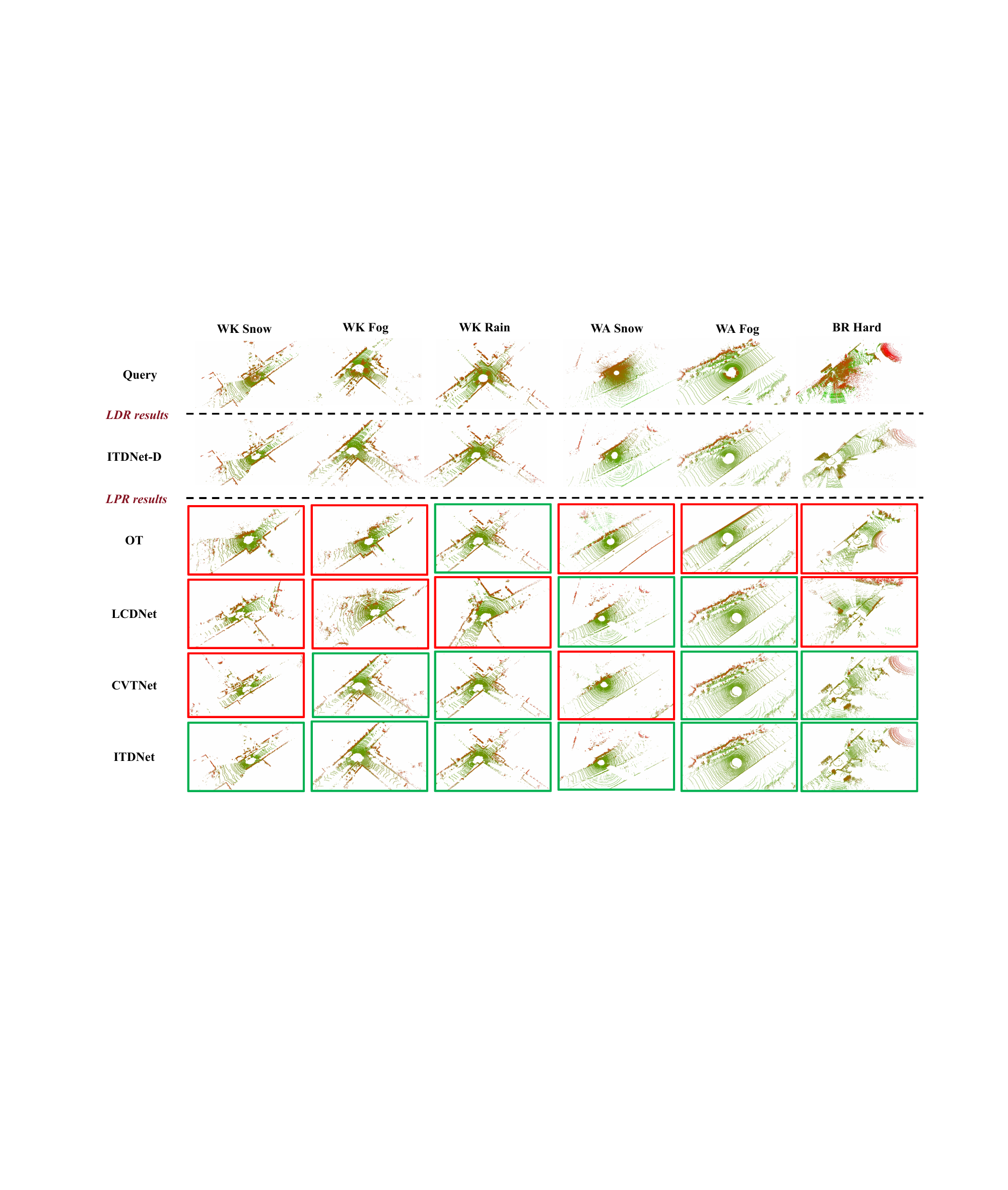}
   \caption{Qualitative results of top-1 retrievals of different LPR methods and the proposed ITDNet-D restoration on the Weather-KITTI, Weather-Apollo, and Boreas datasets. Green boxes indicate correct retrievals, and red boxes denote incorrect ones. The proposed LDR module achieves effective restoration while preserving structural consistency crucial for correct retrievals. Meanwhile, ITDNet attains the best overall LPR performance.}
   \label{lpr_ldr_view}
\end{figure*}

\textbf{Visualization:} 
To enable comprehensive evaluation, we present Recall@$N$ curves ($N \leq 20$) across six representative scenarios from the Weather-KITTI, Weather-Apollo, and Boreas datasets, as shown in Fig.~\ref{prvis}. Our proposed LPR module, ITDNet-P, consistently outperforms standalone baselines across most scenes and retrieval ranges, demonstrating strong discriminative capability. Furthermore, integrating the restoration module ITDNet-D into existing baselines (OT\cite{ot}, LCDNet\cite{LCDNet}, CVTNet\cite{CVTNet}) brings substantial improvements, particularly in the low-$N$ region (Recall@1–5), which is critical for precise localization. Additionally, Fig.~\ref{lpr_ldr_view} visualizes the top-1 retrieval results of different LPR methods as well as the restoration outputs produced by ITDNet-D. The proposed ITDNet-D effectively restores the degraded point clouds, preserving structural consistency crucial for accurate retrievals. Meanwhile, ITDNet achieves the best overall LPR performance across all tested conditions.

\subsection{Evaluation for the LDR task}
\begin{figure}[!t]
	\centering
	\captionsetup[subfigure]{margin=0.8pt} 
	\subfloat[]{\includegraphics[width = 0.5\linewidth]{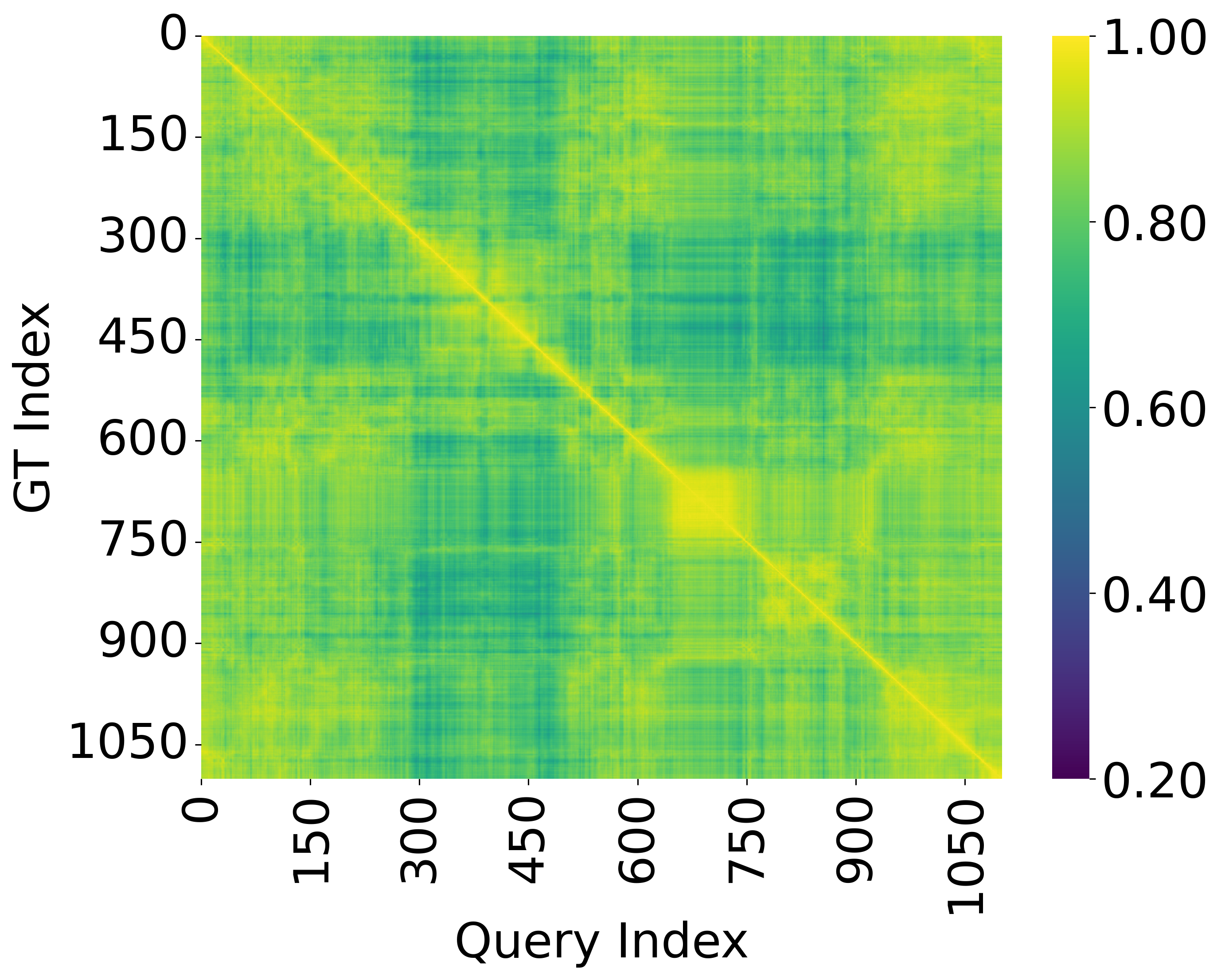}} 
	\hfill
	\subfloat[]{\includegraphics[width = 0.5\linewidth]{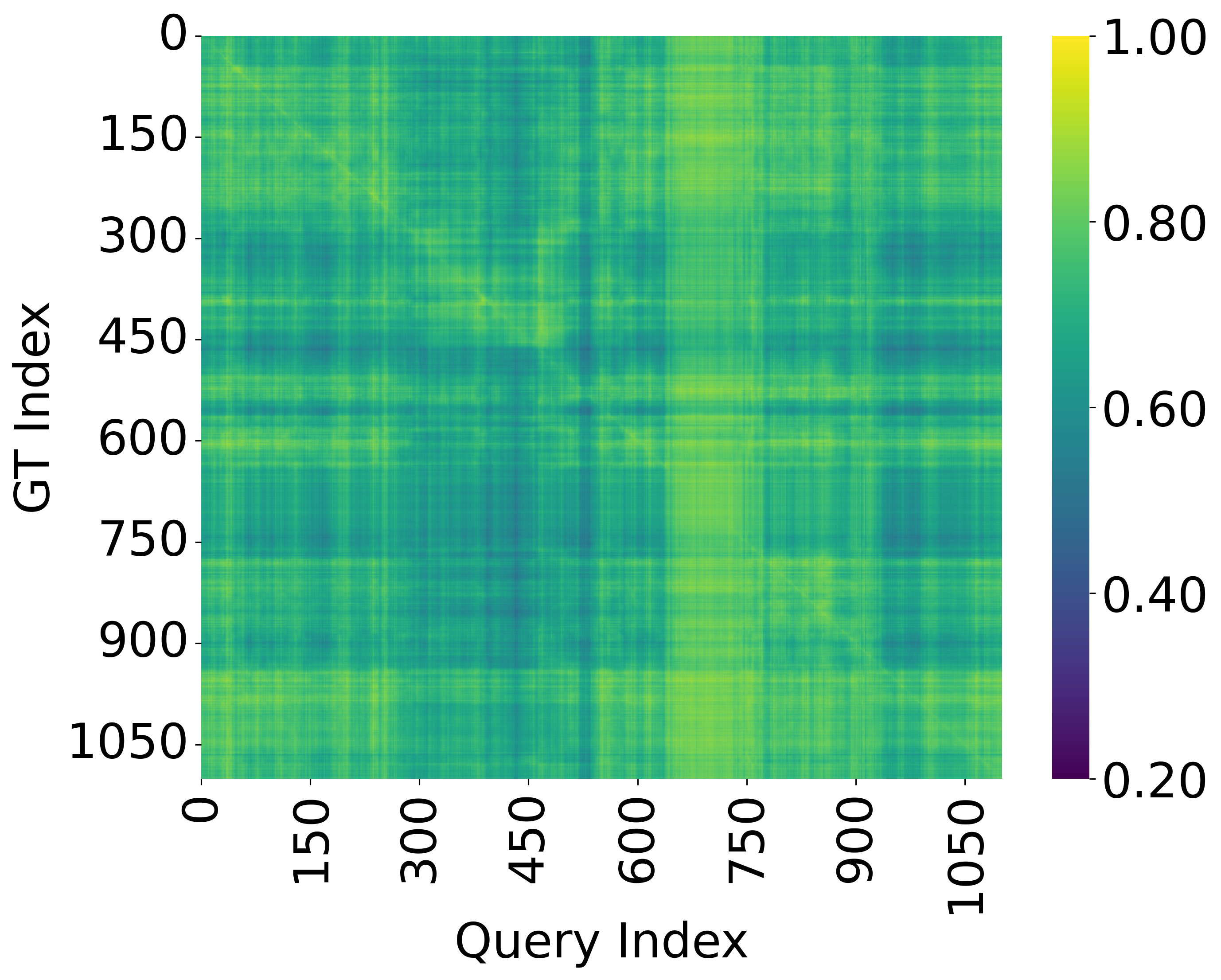}}
	\hfill
    \vspace{-0.4cm}
	\subfloat[]{\includegraphics[width = 0.5\linewidth]{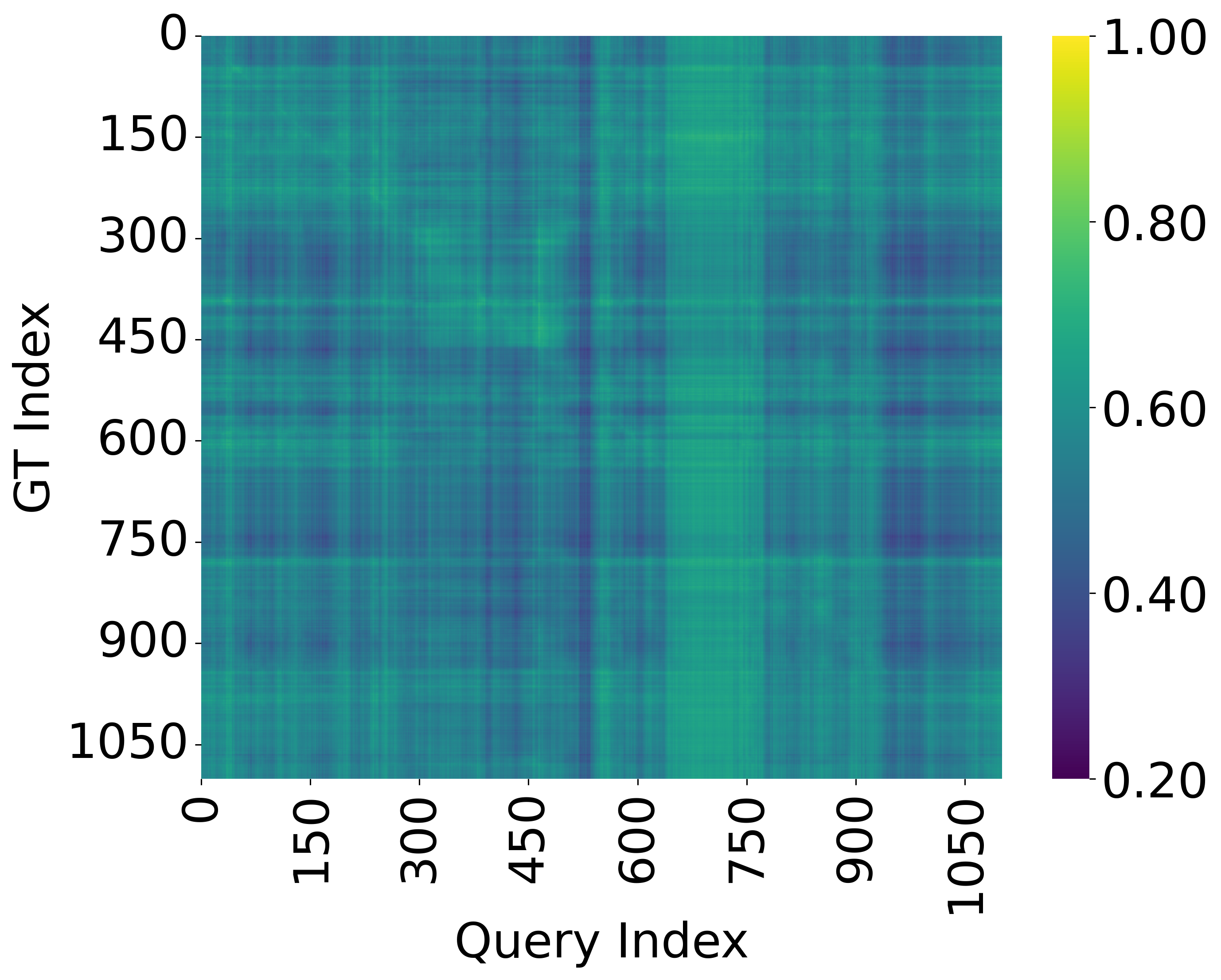}}
	\hfill
	\subfloat[]{\includegraphics[width = 0.5\linewidth]{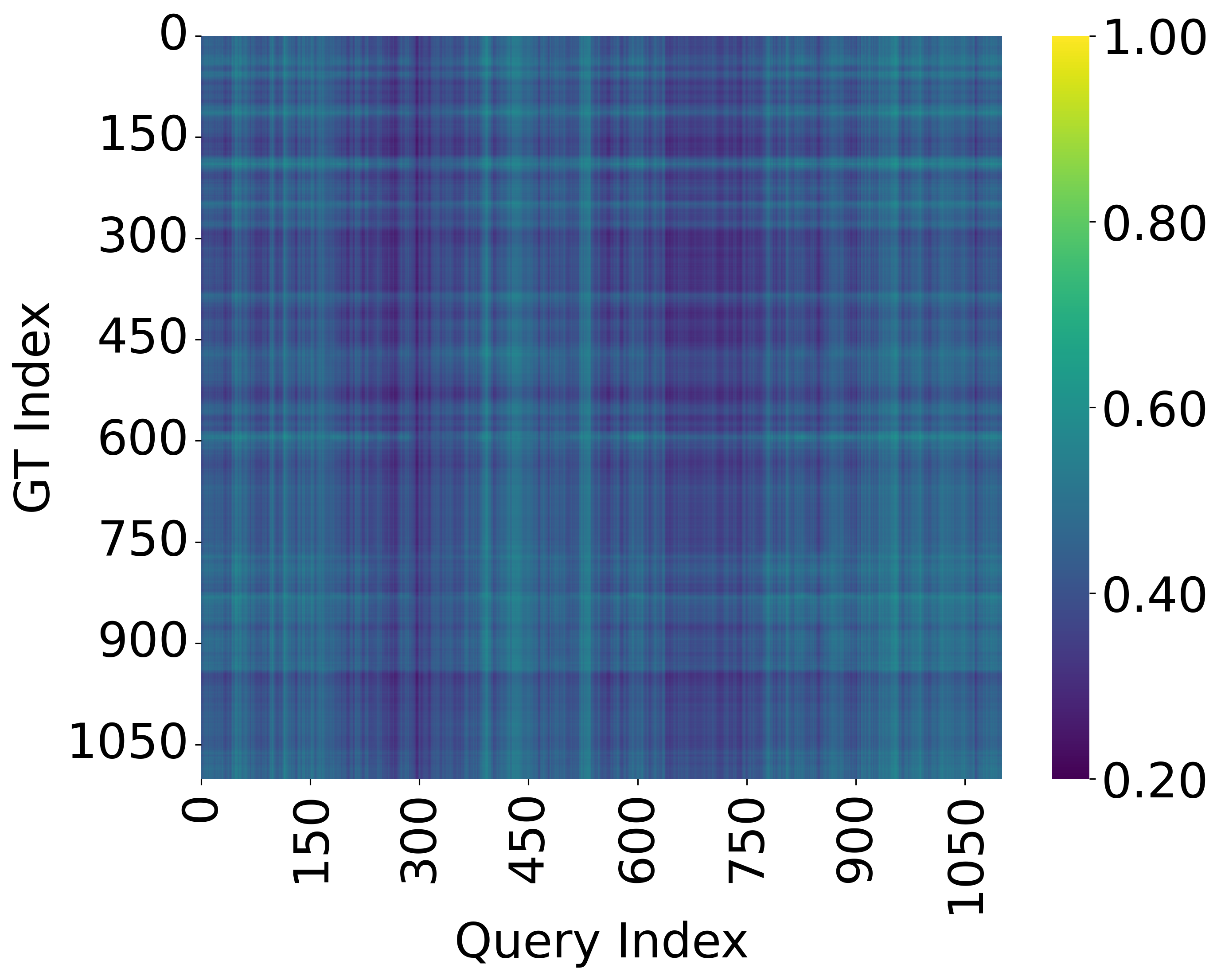}} 
	\caption{ Visualization of feature similarity matrices using the FSS metric for different LiDAR preprocessing methods: (a) ITDNet, (b) ResLPRNet, (c) TripleMixer, and (d) 4DenoiseNet. Our ITDNet restoration shows a clear diagonal pattern, indicating stronger feature consistency.}
	\label{fss}
	\vspace{-0.2cm}
\end{figure}

In this section, we compare the proposed LDR module with three state-of-the-art LiDAR preprocessing methods in terms of their effectiveness for downstream LiDAR place recognition. We evaluate the performance using SSIM (a traditional image-based metric) and our proposed FSS metric. In addition, we report the performance of downstream place recognition using R@1 and R@1\%. For fair comparison, all experiments adopt our ITDNet-P module as the LPR backbone. The baselines include 4DenoiseNet\cite{4denoisenet} and TripleMixer\cite{TripleMixer}, which target LiDAR denoising, and ResLPRNet\cite{reslpr}, a range-image-based restoration approach. Experiments are conducted across three weather conditions in the Weather-KITTI Seq 07 dataset, with average results summarized in Table~\ref{fss_comparison}. Our ITDNet achieves the best performance in FSS (0.95), R@1 (0.83), and R@1\% (0.97), demonstrating strong consistency between restored and clean features. Although ResLPRNet slightly outperforms in SSIM (0.93), its lower FSS and LPR results suggest that pixel-level metrics do not reliably reflect downstream performance. To further validate the FSS metric, we visualize cosine similarity matrices in Fig.~\ref{fss}, where ITDNet exhibits a strong diagonal pattern across Seq 07, confirming high feature alignment and the effectiveness of FSS.

\begin{table}[h]
\centering
\footnotesize
\renewcommand{\arraystretch}{1.2}
\setlength{\tabcolsep}{8pt}
\caption{Comparison of LDR methods on Weather-KITTI}
\begin{tabular}{lcccc}
\toprule
\textbf{Method} & \textbf{FSS↑} & \textbf{SSIM↑}  & \textbf{R@1↑} & \textbf{R@1\%↑}\\
\midrule
4DenoiseNet\cite{4denoisenet} &0.46  &0.56  &0.30 &0.59  \\
TripleMixer\cite{TripleMixer} &0.50  &0.67  &0.37 &0.65 \\
ResLPRNet\cite{reslpr} &0.75  &\textbf{0.93}  &0.66 &0.84  \\
ITDNet &\textbf{0.95}  &{0.91}  &\textbf{0.83} &\textbf{0.98} \\
\bottomrule
\end{tabular}
\label{fss_comparison}
\end{table}

\subsection{Ablation Study} 
We perform an ablation study on ITDNet to assess the impact of each component. Experiments are conducted on Seq 07 of the Weather-KITTI dataset\cite{TripleMixer} and results are averaged across all conditions. We evaluate four key components: DDM and SAG in the LDR module, and MFT and WPN in the LPR module. As shown in Table~\ref{ablation_itr}, the full ITDNet achieves the best performance across all metrics. Removing either DDM or SAG leads to a notable drop, demonstrating the importance of effective LiDAR data restoration for improving downstream place recognition accuracy in adverse weather conditions.

\begin{table}[ht]
\captionsetup{justification=centering, singlelinecheck=false}
\centering
\caption{{Ablation studies on each component of ITDNet}}
\setlength{\tabcolsep}{2pt}
\begin{tabular}{cccccccccc}
\toprule
\multirow{2}{*}{\textbf{Experiment}} & \multirow{2}{*}{\textbf{DDM}} & \multirow{2}{*}{\textbf{SAG}} & \multirow{2}{*}{\textbf{MFT}} & \multirow{2}{*}{\textbf{WPN}} & \multicolumn{4}{c}{\textbf{Metric}} \\
\cmidrule(lr){6-9}
 &  & & & & \textbf{R@1↑} &\textbf{R@5↑}  & \textbf{R@1\%↑} &\textbf{F1↑}  \\
\midrule
\uppercase\expandafter{\romannumeral1} & & & &  &0.64 &0.67  &0.81 &0.73  \\
\uppercase\expandafter{\romannumeral2} & &\checkmark &\checkmark &\checkmark  &0.67 &0.74  &0.85 &0.76    \\
\uppercase\expandafter{\romannumeral3} &\checkmark  & &\checkmark   &\checkmark & 0.69 &0.76  &0.88 &0.77  \\
\uppercase\expandafter{\romannumeral4} & \checkmark &\checkmark &   &\checkmark  &0.77 &0.81  &0.95 &0.83 \\
\uppercase\expandafter{\romannumeral5} & \checkmark &\checkmark &\checkmark & &0.79  &0.84  &0.95 &0.84 \\
\uppercase\expandafter{\romannumeral6} & \checkmark &\checkmark &\checkmark &\checkmark  &\textbf{0.83}  &\textbf{0.91} &\textbf{0.98} &\textbf{0.89} \\
\bottomrule
\end{tabular}
\label{ablation_itr}
\end{table}

\subsection{Runtime} 
To demonstrate the efficiency of ITDNet, we report its runtime performance. The evaluation was conducted on a system equipped with Intel Xeon 8280 CPUs and NVIDIA V100 GPUs, using 4,000 LiDAR scans from our proposed Weather-Apollo dataset. ITDNet contains a total of 29.86 M parameters and achieves an average inference time of 24 ms per scan, demonstrating its capability for real-time operation.

\section{Conclusion and Future Work}
This paper presents ITDNet, an end-to-end framework that jointly performs LiDAR data restoration (LDR) and place recognition (LPR) through an iterative learning strategy to address LiDAR degradation in adverse weather. The LDR module leverages a Dual-Domain Mixer (DDM) and Semantic-Aware Generator (SAG) to recover structural and semantic features, while the LPR module employs a Multi-scale Frequency Transformer (MFT) and Wavelet Pyramid NetVLAD (WPN) to extract robust global descriptors. Experiments on Weather-KITTI, Weather-Apollo, and Boreas demonstrate clear improvements in both restoration quality and recognition accuracy. We also introduce the Feature Similarity Score (FSS), which correlates well with downstream LPR performance. Ablation studies confirm the effectiveness of each component. Despite these advances, limited supervision and the sim-to-real gap remain key challenges. Future work will explore unsupervised learning and more generalizable models for LiDAR perception under adverse weather.

\bibliographystyle{IEEEtran}

\bibliography{IEEEreference}

\end{document}